%% file: main.tex
\documentclass{article} 
\usepackage{iclr2027_conference,times}

\input{math_commands.tex}

\usepackage{hyperref}
\usepackage{xurl}
\usepackage{graphicx, wrapfig}
\usepackage{xspace} 
\usepackage{booktabs,multirow,graphicx}
\newcommand{\name}{PaPaSpec\xspace}
\usepackage[dvipsnames]{xcolor}
\usepackage{caption}  
\definecolor{myblue}{HTML}{5EA6CE}
\definecolor{myorange}{HTML}{E69F00}
\newcommand{\colorsq}[1]{\textcolor{#1}{\rule{1.6ex}{1.6ex}}}

\newcommand{\rebuttal}[1]{\textcolor{Black}{#1}}

\usepackage[most]{tcolorbox}  
\newtcolorbox[auto counter]{takeaway}[1][]{%
  enhanced,  
  breakable,  
  colback=black!5, 
 colframe=black!85,
 boxrule=1.25pt, 
 arc=3pt,
 left=2mm,right=2mm,top=2mm,bottom=1.3mm,  
 before skip=10pt, after skip=10pt,  
 title={Takeaway~\thetcbcounter},  
 colbacktitle=black!85, 
 coltitle=white, 
 fonttitle=\bfseries\small,  
 attach boxed title to top left={yshift=-1.2mm, xshift=2mm},  
 boxed title style={  
 enhanced,  
 arc=3pt,  
 top=0.5mm, bottom=0.5mm, left=1mm, right=1mm,  
 boxrule=0pt, 
 interior engine=empty, 
 },  
 #1 
}

\title{Parallel-Parallel Speculative Decoding}

\author{Michael Hersche$^{1}\thanks{Equal contribution. Research conducted at IBM Research.}$\,\,, Ilias Mc Auliffe$^{1,2 *}$, Nicolas Menet$^{1,2}$, Abbas Rahimi$^{1}$\\
\And
\normalfont $^{1}$IBM Research, $^{2}$Department of Computer Science, ETH Zürich}

\iclrfinalcopy 
\begin{document}

\maketitle

\begin{abstract}
Speculative decoding accelerates memory-bandwidth-bound inference of large autoregressive (AR) target models using a small, fast drafter model.
Target-latent-conditioned parallel drafters, such as DFlash, currently offer the best accuracy-latency tradeoff, leading to the highest reported speedups.
However, latent conditioning still requires executing drafting and verification sequentially, placing drafting latency on the critical path. 
This limits achievable speedups and prevents effective use of emerging hardware that can support concurrent drafting and verification.
We introduce \textbf{pa}rallel-\textbf{pa}rallel \textbf{spec}ulative (PaPaSpec) decoding, which runs a multi-layer latent-conditioned parallel drafter concurrently with verification, hiding drafting latency entirely.
The key enabler is a prefix and depth-causal latent conditioning scheme that restricts each drafter layer to only access the target latents computed so far.
Hiding drafting behind verification not only removes it from the critical latency path, but also buys the drafter additional time to refine its draft.
We exploit this by looping over intermediate layers until new conditioning information arrives, which increases the acceptance length while keeping the drafter size constant.
\rebuttal{For single-request decoding on H100 GPUs, PaPaSpec accelerates AR generation across seven benchmarks (covering math, coding, and chat) by 3.44--5.98$\times$ on average, a 19--31\% improvement over DFlash and 9--18\% over D-PACE.
The gain transfers to vLLM in single-batch (5--10\% over D-PACE) and multi-batch servings of up to 16 concurrent requests (1.7--11.1\% on SPEED-bench). 
}
\end{abstract}

\section{Introduction}

Autoregressive (AR) generation~\citep{vaswani_attention_2017, brown_language_2020} remains the dominant generation paradigm in large language models (LLMs).
While AR training can be efficiently parallelized via teacher forcing~\citep{williams1989learning}, AR inference is inherently sequential, generating tokens one at a time.
Predicting a single token requires loading the model parameters and the key-value (KV) cache from off-chip memory to the on-chip compute units.
In low-concurrency serving, where only a few requests are processed at once, AR generation is hence often memory-bandwidth-bound, leaving spare compute unutilized.
As a viable solution, speculative decoding~\citep{leviathan_fast_2023, chen_accelerating_2023} accelerates inference by leveraging this spare compute to verify, in parallel, a sequence of possible next tokens proposed by a cheaper drafter model.\looseness-1

\begin{figure}[t]
    \centering
    \includegraphics[width=1.0\linewidth]{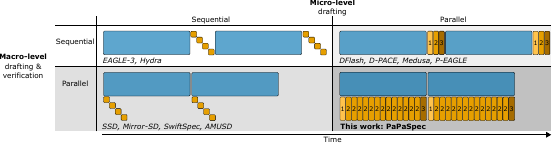}
    \caption{Verification (\colorsq{myblue}) and drafting (\colorsq{myorange}) timeline for different approaches. 
    Performing verification and drafting macro-sequentially puts drafting into the critical latency path, whether the drafter itself is micro-level sequential or parallel. 
    Instead, performing verification and drafting in macro-level parallel hides the drafting latency, but so far only supports micro-level sequential drafters. 
    This work presents \name, which performs macro-level parallel verification with micro-level parallel drafting, hiding the drafting latency while further exploiting spare compute via looping.
    }
    \label{fig:concept}
\end{figure}

While effective, speculative decoding faces two distinct forms of sequentiality that negatively impact the achievable decoding speedup (see Figure~\ref{fig:concept}). 
First, drafting itself is often sequential, generating candidate tokens autoregressively one token after another. 
We refer to this as \emph{micro-level} sequentiality. 
Second, drafting and verification are typically executed synchronously, such that verification must complete before the next drafting can begin. 
We refer to this as \emph{macro-level} sequentiality. 
Eliminating both forms of sequentiality is essential for maximizing speculative decoding speedups. 

Recent progress in resolving micro-level sequentiality is largely driven by conditioning drafters on the target model's latent states. Because these latents encode information about future tokens~\citep{samragh_your_2025}, they allow the drafter to predict multiple tokens in parallel.
While target-model conditioning improves the acceptance rates of sequential drafters~\citep{li_eagle_2024, li_eagle2_2024, li_eagle3_2025, ankner_hydra_2024}, it acts as a critical enabler for parallel drafters~\citep{stern_blockwise_2018, cai_medusa_2024}, with conditioning depth ranging from a single hidden state to multiple target layers.
DFlash~\citep{chen_dflash_2026} pushes this paradigm even further: it conditions on the accumulated \textit{history} of projected multi-layer target latents rather than just the last-token hidden states.
While such parallel drafting, optionally enhanced with improved training objectives~\citep{wu_dpace_2026} or light-weight AR correction~\citep{cheng_dspark_2026, huang_domino_2026a}, achieves state-of-the-art (SoTA) speedups, the depth of the drafter, and hence its acceptance length, is limited due to being on the critical latency path.

Concurrently, another line of work has sought to eliminate macro-level sequentiality by executing drafting and verification asynchronously in parallel~\citep{mcdanel_amusd_2025,kumar_speculative_2026}.
In fact, dedicated accelerators for speculative drafting are increasingly available~\citep{bartlett_corsair_2026, bhendawade_mirror_2025}, such as the recently announced Groq 3 LPX on the NVIDIA Vera Rubin platform~\citep{weidman_groq3lpx_2026}. 
These devices allow launching drafting before the target model's forward pass completes, thereby hiding drafting latency behind verification.
However, current macro-parallel approaches rely on standard, decoupled sequential drafters rather than tightly conditioned parallel ones: multi-layer latent conditioning as used by SoTA micro-parallel drafters requires target latents, which are not yet available when drafting occurs concurrently with verification.

In this work, we propose \emph{\textbf{pa}rallel-\textbf{pa}rallel \textbf{spec}ulative (\name) decoding}, which removes both levels of sequentiality: a tightly conditioned parallel drafter (micro-level) operates in parallel with the target model (macro-level), hiding drafting behind verification (see Figure~\ref{fig:concept}).
The key innovation is a new latent conditioning that attends only to already computed target model layers and prefix, making conditioning available incrementally as the target verifies.
Hiding drafting behind verification goes beyond merely removing the drafting latency: it buys the drafter time, which we can spend on additional compute by increasing the drafter's depth.
We show that looping internal layers in the drafter can effectively increase acceptance length while keeping the number of drafter parameters constant.
This work makes the following contributions:
\begin{itemize}
\item We propose prefix and depth-causal conditioning for parallel drafters, where each drafter's layer only accesses the target model latents that have already been computed during verification. This effectively removes the requirement in prior target-conditioned drafters (e.g., DFlash) that the target model's forward pass complete before drafting can begin.
\item We introduce looping to parallel drafting, where intermediate layers are looped to increase acceptance length while keeping the number of parameters constant: a single drafter can thus adapt to varying acceptance length, compute, and latency demands at inference time.
\item Combining our prefix and depth-causal conditioning, looping, and macro-parallelism yields \name{}, a parallel drafter that drafts in parallel with the target model verification, combining micro- and macro-level parallelism in speculative decoding.
\item \rebuttal{We evaluate \name on three target models and seven benchmarks on H100 GPUs. \name achieves speedups of 3.44--5.98$\times$ over AR, improving by 19--31\% over DFlash and 9--18\% over D-PACE, with consistent gains on vLLM (13--24\% and 5--10\%). Even when drafting on the host CPU, \name still improves over D-PACE by 3.1\%.}
\item \rebuttal{We extend \name to concurrent serving by drafting without a bonus token, which removes cache misses, together with a Markov correction head. On Gemma-4-26B-A4B-it, this improves the speedup over D-PACE by 1.7--11.1\% for up to 16 concurrent requests.}
\end{itemize}

\section{Preliminaries}
\label{sec:prelimiaries}

\subsection{Speculative decoding}
Speculative decoding accelerates AR inference of a large target model $p$ by using a faster draft model $q$. 
Given a prefix $x_{1:n}:= (x_1, x_2, ..., x_n)$ of length $n$, where each token $x_i$ is in the vocabulary $\mathcal{V}$, the drafter proposes a continuation of $\gamma$ tokens $\hat{x}_{n+1:n+\gamma} \sim q(\cdot | x_{1:n})$. 
The target model then verifies the entire draft in one pass. 
In a memory-bandwidth-bound regime, the verification/target latency $t_T$ is largely insensitive to the drafting length, i.e., is assumed to equal the target model's per-token AR latency. 
The drafted tokens are verified from left to right, either via exact-match acceptance in the greedy setting~\citep{stern_blockwise_2018}, or via acceptance sampling in the stochastic setting~\citep{leviathan_fast_2023}.
When a token is rejected, a bonus token is sampled from the normalized residual distribution between the target and drafter.
This approach is lossless with respect to the target distribution. 
We denote the average acceptance length (including the bonus token) as $\tau\in [1, \gamma+1]$.

\begin{figure}[b]
    \centering
    \includegraphics[width=0.9\linewidth]{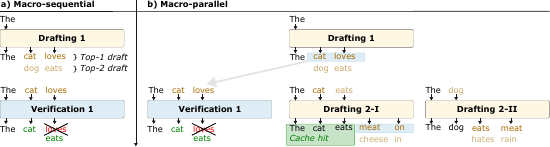}
    \caption{\textbf{(a) Macro-sequential decoding.} Drafting is followed by verification, where the top-1 draft (``cat loves'') is verified. The target model accepts the first token (``cat''), while rejecting the second (``loves'') and resampling a bonus token (``eats''). 
    \textbf{(b) Macro-parallel decoding.} While the target model verifies the top-1 draft, the drafter speculates on two possible next prefixes (``The cat eats'' and ``The dog''), and generates drafts concurrently. If the verified sequence matches one of the speculated prefixes, the corresponding draft is already available and can be verified next. 
    }
    \label{fig:ssd}
\end{figure}

\paragraph{Macro-sequential verification and drafting} 
As shown in Figure~\ref{fig:ssd}a, the standard speculative decoding framework issues drafting and verification sequentially. 
The expected speedup is $\eta=\frac{\tau \cdot t_T}{t_D + t_T}$, 
where $t_T$ is the target model latency and $t_D$ the drafting latency. 
Since the drafter is on the critical latency path, naïvely aiming for acceptance length improvements by scaling up the drafter, which may increase drafting latency, does not necessarily translate into a wall-clock speedup improvement.

\paragraph{Macro-parallel verification and drafting with SSD}
One viable approach is to hide the drafting latency behind the verification, as proposed by \citet{kumar_speculative_2026} in speculative speculative decoding (SSD). 
Figure~\ref{fig:ssd}b illustrates this concept. 
Here, the next drafting starts while the previous draft is still being verified. 
The main challenge is that the result of the current verification (the accepted length and the bonus token) is not yet known. 
To resolve this, the next drafting round is based on a \textit{speculated prefix}, which consists of the first $\hat \tau -1$ tokens of the verification sequence, followed by a bonus token that differs from the corresponding token in the submitted verification sequence (e.g., the top-2 token from the previous draft). Here, $\hat \tau \in [1, \gamma+1]$. 
To increase the chance of speculating the correct prefix, a \emph{cache} containing multiple prefixes is used to draft in parallel.
If the verification outcome matches one of the speculated prefixes (cache hit), the corresponding draft can be verified without any additional latency. 
If none of the speculated prefixes match or the entire draft was accepted, an additional sequential draft has to be performed, which again falls on the critical latency path.
Hence, even though predominantly hidden, drafting latency still plays a role in SSD. 

\subsection{Micro-parallel drafting with DFlash}
In contrast to SSD, which parallelizes on the system level, micro-parallel drafting parallelizes the drafting step itself, generating all draft tokens in a single forward pass.
DFlash~\citep{chen_dflash_2026} realizes this with a single-step masked diffusion model~\citep{austin_structured_2021, arriola_block_2025a}, denoising all $\gamma$ positions simultaneously and yielding a drafting latency independent of $\gamma$.
Key to its accuracy is conditioning the drafter on the target model's latents from the previous verification step.

Recall that during verification, the target model computes $p\!\left(x_{n+1} \mid x_{1:n}\right)
  = \big(\mathrm{LM_{head}} \circ
     \mathcal{T}^{(L_T)} \circ \cdots \circ \mathcal{T}^{(1)} \circ
      \mathrm{E}\big) (x_{1:n})$,
where $\circ$ denotes function composition, $\mathrm{E}(\cdot)$ the input embedding, $\mathcal{T}^{(l)}(\cdot)$ the $l$-th Transformer decoder layer, and  $\mathrm{LM_{head}}(\cdot)$ the projection (including softmax) from the model's final latents to the simplex over the vocabulary. 
In addition, the target also provides the latents
\begin{equation*}
 h_{1:n}^{(l)} = \big(\mathcal{T}^{(l)} \circ \cdots \circ \mathcal{T}^{(1)} \circ \mathrm{E}\big)\big(x_{1:n}\big) \in \R^{n \times d} \text{ for all } l \in [1, L_T].    
\end{equation*}
As $\mathcal{T}^{(l)}$ uses causal attention, all latents from the previous verifications (e.g., $h^{(l)}_{1:n-1}$) can be cached.
DFlash conditions on a subset $\{l_1, \cdots l_m\}$ of the target latents via concatenation and projection:  
\begin{equation}
\label{bidi-condition}
  \textstyle c_{1:n} = W \big[h_{1:n}^{(l_1)} \Vert \cdots \Vert h_{1:n}^{(l_m)}\big]
      = \sum_{i=1}^{m} W^{(i)} h_{1:n}^{(l_i)}, 
\end{equation}
where  $c_{1:n} \in \R^{n \times d}$ and $W^{(i)} \in \R^{d\times d}$. 
These projected latents act as a conditioning signal via injection into the drafter's attention. 
Additionally, the input to the drafter is a bonus token $x_{n+1}$ and $\gamma$ mask tokens $z = \mathrm{E}\big(x_{n+1}, [M], \dots, [M]\big) \in \R^{(\gamma+1)\times d}$. 
The drafter thus computes the marginals
\begin{equation*}
  q\big(x_{n+2:n+\gamma+1} \mid x_{1:n+1}\big)
  = \big(\mathrm{LM_{head}} \circ
      \mathcal{D}^{(L_D)}(\,\cdot\,;c_{1:n}) \circ \cdots \circ \mathcal{D}^{(1)}(\,\cdot\,;c_{1:n})\big)\big(z\big),
\end{equation*}
where $L_D$ is the number of layers and $\mathcal{D}^{(l)}(\cdot ; c_{1:n})$ is a Transformer layer with bidirectional attention in the positions of $z$ and causal attention to RMSNorm-scaled $c_{1:n}$.  
$\mathrm{E}$ and $\mathrm{LM_{head}}$ are borrowed from the target (i.e., parameter-sharing). 
Two elements prohibit straightforward macro-parallel decoding in DFlash. Firstly, the drafter needs the bonus token $x_{n+1}$, which again requires speculation as in SSD. 
Second, and more detrimentally, the conditioning vector $c_{1:n}$ depends on all the chosen verifier hidden states, some of which might only be available towards the end of verification.

\section{Parallel-parallel speculative decoding (\name)}
\begin{figure}[t]
    \centering
    \includegraphics[width=1\linewidth]{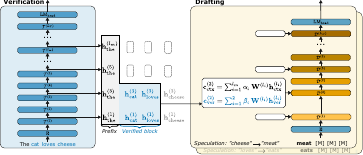}
    \caption{\name architecture. 
    The target model verifies the previous draft (e.g., ``cat loves cheese''), while the drafter concurrently generates drafts based on speculated acceptance lengths and bonus tokens (e.g., ``meat'' or ``eats'').  
    The drafter is conditioned on layer-specific latents following a prefix and depth-causal scheme in which each drafter layer only accesses target latents computed so far.
    Depending on verification progress, latents from all layers ($h_\text{the}$) or only the so far completed layers ($h_\text{cat}$ and $h_\text{loves}$) enter the conditioning. 
    Finally, the drafter can leverage the spare time before the next latents arrive to increase drafting quality by looping over intermediate layers. 
    }
    \label{fig:architecture}
\end{figure}

This section presents \name{}, a tightly target-latent-conditioned micro-parallel drafter that operates in a macro-parallel speculative decoding framework, illustrated in Figure~\ref{fig:architecture}. 
We first introduce our prefix and depth-causal conditioning that removes the dependency on fully completed target verification, enabling concurrent drafting and verification. 
We then show how the gained spare latency can be converted into improved draft quality through layer-wise looping. 
Finally, we describe the integration into SSD and present a novel dynamic budget-allocation strategy.

\subsection{Prefix and depth-causal conditioning}
DFlash natively requires the availability of all selected target hidden layers.
Consequently, the conditioning signal $c_{1:n}$ becomes available only after the final (selected) target layer has been computed.
Hence, naïvely applying macro-parallel SSD with such conditioning would hardly provide any wall-clock time improvement, as drafting would still depend on, and wait for, the completion of verification.
To remove this bottleneck, we break the depth-bidirectional dependency between target and drafter by instead introducing a \emph{prefix and depth-causal conditioning mechanism}.

\paragraph{Architecture}  
Instead of using the same conditioning for every drafter layer, we follow \citet{zhang_dflare_2026} and modify the DFlash architecture with layer-specific conditioning. Our modification differs in that we retain the projection matrices $W^{(i)}$. 
More decisively, we treat previously verified tokens and currently verified tokens differently.
For tokens belonging to completed verification rounds (i.e., the prefix), all verifier layers are available and full depth-bidirectional conditioning can be applied.
For tokens in the block currently undergoing verification, only a subset of verifier layers has been computed; therefore, conditioning can only access latents completed so far. 
For a token index $t$, assuming tokens up to position $n$ have already been  verified and speculating that we will have $\hat \tau$ tokens accepted, we define the prefix and depth-causal conditioning for drafter layer $l$ as
\begin{equation*}
  c^{(l)}_{t} =
  \left\{
\begin{array}{@{}l@{\qquad}l@{\qquad}l@{}}
    \displaystyle\sum\nolimits_{i=1}^{m} \rebuttal{\alpha_{i}}\, W^{(i)} h_{t}^{(l_i)},
      & 1 \le t \le n,
      & \text{(prefix)} \\[1ex]
    \displaystyle\sum\nolimits_{i=1}^{l} \rebuttal{\beta_{i}}\, W^{(i)} h_{t}^{(l_i)},
      & n < t \le n + \hat \tau -1
      & \text{(tokens under verification)},
  \end{array}
  \right.
\end{equation*}
\rebuttal{where $\alpha, \beta \in \mathbb{R}^{m}$ are learned mixing weights shared by
all drafter layers. Draft layer $l$ differs only in which latents it may read
(details in Listing~\ref{lst:ssd-conditioning}). Branch $(\hat{\tau}, b)$ is masked to condition only on positions $t \leq n+\hat{\tau}-1$. A cache hit thus implies no mismatch between drafting and verification.
}

\paragraph{Training procedure} 
By design of SSD, the drafter operates under varying acceptance lengths $\hat \tau \in [1, \gamma+1]$ of the previous block. Since we only condition depth-causally to unverified token positions, there is a shift in the conditioning distribution depending on the value of $\hat \tau$. We model the acceptance length as a geometric distribution under the assumption of i.i.d. acceptance probability $\rho$ per position~\citep{leviathan_fast_2023}. To cover all $\hat \tau$ during training, we thus sample $\hat \tau$ from $\Pr(\hat \tau{=}k) = (1-\rho)\,\rho^{k-1}$ and truncate it to $[1, \gamma]$ to match the bonus token availability at inference.

\subsection{Looped parallel drafting}
Because \name removes drafting from the critical latency path, drafting can consume up to the same latency budget as verification without increasing the wall-clock time. 
A natural way to improve the draft quality of masked diffusion models would be to increase the number of denoising steps. 
However, multi-step denoising requires access to the full verifier conditioning signals and therefore would reintroduce the synchronization constraints between drafting and verification.
Instead, we increase the effective depth of the drafter by reusing intermediate layers, a technique known as looping, which has been shown to achieve depth-scaling behavior similar to multi-step diffusion~\citep{lee_looped_2026, carballo-castro_scaling_2026}.
We repeatedly apply the intermediate layers of the drafter, e.g., the layers $\mathcal{D}^{(2)}$, $\mathcal{D}^{(3)}$, and $\mathcal{D}^{(4)}$ in a 5-layer drafter \rebuttal{as a block~\citep{geiping_scaling_2025}.}

Looping allows the drafting compute to be scaled dynamically, for example based on the available drafter compute budget, and does not increase the number of parameters.
In particular, the drafter can still be executed without any loops, which is important for latency-sensitive drafting in case of an SSD cache miss.
Unlike the original SSD framework, which instantiates both an accurate, high-latency primary drafter and a cheap, low-latency backup drafter, \name covers both modes with a \textit{single model}. 
Accordingly, we train the drafter to operate reliably at both compute budgets:
for every training batch, we sample the number of loops as $r = 1 + b(R-1)$, with $b \sim \mathrm{Bernoulli}(0.5)$, where $R$ is the maximum number of loops.

\subsection{Integration into speculative speculative decoding (SSD)}
Figure~\ref{fig:architecture} illustrates the adoption of \name into SSD. 
During verification, the drafter concurrently generates drafts for the next SSD iteration. 
These drafts are conditioned on all of the prefix latents and incrementally available depth-causal latents of the currently verified block, as well as the speculated bonus tokens sampled from the previous drafting round.
In the event of an SSD cache miss, drafting falls back to macro-sequential execution, disabling looping to minimize drafting latency. 

\paragraph{Confidence-based budget allocation}
In SSD, the drafter generates candidate continuations from a set of speculated pairs (acceptance length, bonus token), with the bonus token derived from the drafter.
A key design choice is how many bonus tokens $f_k$ should be considered per speculated accept length $k$, given a fixed drafting budget $B=\sum_{k=1}^{\gamma+1} f_k$.
Prior work used fixed allocation strategies, such as uniform or geometric distributions~\citep{kumar_speculative_2026}. 
However, the confidence of a draft has been shown to correlate with the acceptance length~\citep{li_eagle2_2024, wu_dpace_2026}.
We therefore propose a confidence-based budget allocation scheme that dynamically distributes the available drafting budget according to the drafter's predicted token distribution $q_i$. 
More precisely, from the set $\mathcal{C}=\{k, b\}$ of all possible initial positions $k$ and bonus tokens $b$, we select the subset 
\begin{equation*}
\textstyle\mathcal{C}^\star = \operatorname{top-B}_{(k,b)\in\mathcal{C}}\, \sigma(k,b) \quad \text{with} \quad \sigma(k,b) = \log q_{k}(b) + \sum_{i=1}^{k-1} \log q_i(\hat{x}_i). 
\end{equation*}
I.e., we select the branches with maximal log-probability $\sigma(k,b)$ under the drafter's marginals, a surrogate for the verifier.
The induced drafting budget per position $k$ is therefore $f_k = |{b: (k,b) \in \mathcal{C}^\star}|$. 
When the drafter is uncertain about its early predictions, more budget will be allocated towards the start of the block. Likewise, if the drafter is highly confident in early positions but uncertain around a mid point, the budget will be concentrated around that point.

\paragraph{Multi-device deployment}
By design, \name{} executes the verifier and drafter concurrently on separate compute devices.
In our current implementation, we deploy the target model and drafter on two H100 GPUs. 
The primary additional cost introduced by \name{} is the transfer of verifier latents to the drafter.
For a drafted block of length $\gamma$ currently under verification, the target model transmits the selected latents from the $m$ layers with dimensionality $d$. 
This amounts to a total communication volume of $m \cdot \gamma \cdot d$ values. 
For a practical configuration with $m=5$, $\gamma=16$, $d=4096$, and BF16 precision, the communication overhead is only 655\,kB per verification. 
Compared to the cost of executing the verifier, this memory-bandwidth overhead is negligible.
On NVIDIA GPUs, the execution of the drafter and verifier layers does not have to be synchronized, e.g., by using the NCCL LL~\citep{zhang_swiftspec_2026}.
In our implementation, both models communicate over non-blocking point-to-point NCCL and are executed through CUDA Graph replays using the HuggingFace Transformers \citep{wolf2019huggingface} backend with \texttt{torch.compile} (see Appendix~\ref{appendix:verifier-drafter-communication}).
\rebuttal{
While we mainly evaluate \name{} using a two-GPU setup, this deployment model naturally aligns with emerging accelerator architectures that provide dedicated hardware for draft generation, such as the d-Matrix Corsair~\citep{bartlett_corsair_2026}, the NPU in Apple's M-series~\citep{bhendawade_mirror_2025}, or the Groq 3 LPX on the NVIDIA Vera Rubin~\citep{weidman_groq3lpx_2026}. In fact, we demonstrate a use case of such asymmetric verification-draft compute by drafting on the CPU in Section~\ref{sec:res-serving}. 
}

\section{Experiments}

\subsection{Setup}
\paragraph{Target models and evaluations} 
We evaluate on Qwen3-4B and Qwen3-8B~\citep{yang_qwen3_2025}, as well as Gemma-4-26B-A4B-it~\citep{gemmateam2026gemma4}.
Our ablations focus on Qwen3-8B. 
We evaluate on a diverse range of benchmarks across three domains. 
For \textit{math reasoning}, we use GSM8K~\citep{mirzadeh_gsmsymbolic_2025}, MATH~\citep{lightmanLetsVerifyStep2024}, and AIME26 through MathArena~\citep{dekoninck2026matharena}; 
for \textit{coding}, we use HumanEval \citep{chen_evaluating_2021}, MBPP~\citep{austin2021program}, and LiveCodeBench (v1-v6)~\citep{jain2025livecodebench};
and for \textit{chat}, we use MT-bench \citep{zheng2023judging}. 
For each method, we report the speedup with respect to standard AR decoding. We benchmark the macro-sequential drafters DFlash and D-PACE on a single H100, while \name{} uses two (one each for the target model and drafter). 
\rebuttal{We report results on two backends: Hugging Face Transformers (with \texttt{torch.compile} and CUDA Graph replays) and vLLM (version 0.31).}
%
\paragraph{Drafter details} 
Following DFlash, we train our \name with a block size of $\gamma=16$ and $R=3$ loops using a mixture of 800k samples from NVIDIA's Nemotron Post-Training Dataset V2~\citep{NemotronPostTrainingDatasetV2} and CodeAlpaca~\citep{luo2024wizardcoder}. 
For our conditioning ablations, we used a subset of 100k samples.
%
%
We use the D-PACE~\citep{wu_dpace_2026} loss and rely, except for depth-causal conditioning and looping, on the exact same architectural configuration as DFlash. 
\begin{wrapfigure}{r}{0.45\linewidth}
    \centering
    \vspace{3pt}
    \includegraphics[width=\linewidth]{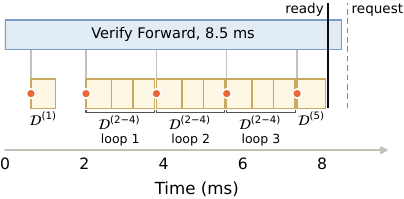}
    \vspace{-10pt}
    \caption{\rebuttal{\name{} timeline for one macro-parallel decoding step with Qwen3-8B, three loops, and a cache budget of 32.}}
    \label{fig:ssd_timeline}
    \vspace{-20pt}
\end{wrapfigure}
Besides testing against publicly released DFlash checkpoints, we train an additional DFlash drafter with the D-PACE loss as a stronger baseline. 
Appendix~\ref{appendix:experimental-details} provides more details.\looseness-1

\subsection{Main results}
\input{tables/main_table_compiled}

Table~\ref{tab:main} reports our main results. 
On Qwen3-8B in greedy generation, \name reaches an average speedup of \rebuttal{5.98}$\times$, which is \rebuttal{18}\% higher than our D-PACE baseline and \rebuttal{31}\% higher than DFlash. 
The advantage transfers to Qwen3-4B with a \rebuttal{5.64}$\times$ speedup and \rebuttal{31}\% gain over DFlash, as well as to Gemma-4-26B-A4B-it, which features an overall lower speedup of \rebuttal{3.81}$\times$ but still provides a \rebuttal{25\%} gain over DFlash. 
When the target models sample with a temperature of $1$, the overall speedups decrease, but we maintain our relative advantage over DFlash (\rebuttal{19--27\%}). 
Figure \ref{fig:ssd_timeline} depicts timing measurements of \name for one decoding step with Qwen3-8B. 
The verification takes 8.5\,ms, whereas a single loop drafter layer takes \rebuttal{0.62}\,ms and a layer looped 3 times takes \rebuttal{1.73}\,ms. As we can see, the next draft tokens are available \textit{before} the target finishes the verification; hence, the drafting latency is hidden.
See Appendix~\ref{appendix:ssd-timeline-appendix} for a detailed timing analysis.

\subsection{\rebuttal{Serving}}\label{sec:res-serving}

\paragraph{vLLM single-concurrency.}
\input{tables/vLLM_main}

\rebuttal{
Table~\ref{tab:vllm_serving} shows the vLLM deployment results. 
For Qwen3-8B on H100s, \name decodes 10\% faster than D-PACE. On MATH500, this corresponds to up to 1153 tok/s, compared to 1046 tok/s for D-PACE. The gains over both baselines persist for the Qwen3-4B and the MoE Gemma-4-26B-A4B. 
Table~\ref{tab:serving_stacks} in Appendix~\ref{appendix:vllm} provides more detailed results.
}

\paragraph{vLLM concurrent serving.}
\begin{wraptable}{r}{0.65\linewidth}
    \centering
    \vspace{-12pt}
    \caption{\rebuttal{Concurrent serving on Gemma-4-26B-A4B-it with vLLM (SPEED-Bench throughput split~\citep{abramovich_speedbench_2026}, greedy, H100).
    D-PACE and \name{} draft $\gamma$=8 tokens; \name{} drafts anchor-free with a Markov correction head.}}
    \label{tab:serving_main}
    \newcommand{\up}[1]{\textcolor{green!50!black}{+#1\%}}
    \newcommand{\down}[1]{\textcolor{red!70!black}{$-$#1\%}}
    \setlength{\tabcolsep}{2pt}
    \resizebox{\linewidth}{!}{%
    \begin{tabular}{ll rrrrrrr}
\toprule
\multirow{2}{*}{\bf Input} & \multirow{2}{*}{\bf Method} & \multicolumn{7}{c}{Output tok/s at concurrency $c$} \\
\cmidrule(lr){3-9}
 & & 1 & 2 & 4 & 8 & 16 & 32 & 64 \\
\midrule
\multirow{4}{*}{1k}
  & AR        & 174 & 280 & 441 & 702 & 1083 & 1597 & 2365 \\
  & D-PACE    & 284 & 459 & 739 & 1198 & 1920 & \textbf{3113} & 4132 \\
  & \name{}   & \textbf{308} & \textbf{487} & \textbf{769} & \textbf{1233} & \textbf{1953} & 3066 & \textbf{4175} \\
\cmidrule(lr){2-9}
  & \name gain      & \up{8.2} & \up{6.2} & \up{4.1} & \up{2.9} & \up{1.7} & \down{1.5} & \up{1.0} \\
\midrule
\multirow{4}{*}{2k}
  & AR        & 164 & 278 & 426 & 694 & 1033 & 1562 & 2315 \\
  & D-PACE    & 272 & 445 & 704 & 1130 & 1868 & \textbf{2826} & 3653 \\
  & \name{}   & \textbf{302} & \textbf{480} & \textbf{763} & \textbf{1219} & \textbf{1918} & 2761 & \textbf{3750} \\
\cmidrule(lr){2-9}
  & \name gain     & \up{11.1} & \up{8.0} & \up{8.3} & \up{7.9} & \up{2.7} & \down{2.3} & \up{2.7} \\
\bottomrule
\end{tabular}
    }
\end{wraptable}
\rebuttal{
We extend \name{} to concurrent serving in vLLM.
Since a single cache miss stalls the entire batch, the probability of a successful macro-parallel step decays exponentially with concurrent requests $c$.
We therefore adapt \name{} to also draft \textit{without} an anchor token, using a placeholder $[P]$ with a trainable embedding in its place, guaranteeing a 100\% cache hit rate.
A lightweight correction head~\citep{cheng_dspark_2026} then connects the eventual bonus token with the anchor-free draft.
The adapted \name{} can still draft with exact anchor tokens, which we exploit by augmenting the SSD cache with exact-anchor drafts.
Table~\ref{tab:serving_main} shows that this variant improves the throughput of Gemma-4-26B-A4B-it over both AR decoding and our D-PACE baseline.
At higher concurrency ($c \geq 32$), the drafting cost, which grows linearly with $c$, becomes compute-bound and can no longer be consistently hidden behind verification.
Appendix~\ref{appendix:batched} provides the full setup and more results.
}

\paragraph{Macro-parallel drafting on the CPU.}
\rebuttal{
Finally, we consider a deployment in which the macro-parallel drafting is offloaded to the host CPU, which typically remains underutilized in serving~\citep{mo_cpu_gpu_2026}.
We run the drafter in a custom C++ implementation on the CPU's AMX units, with linear layers quantized to W8A8.
Only the last drafter layer and the LM head run on the verifier GPU, concurrently with the final target layers.
Due to lower compute availability, we reduce the cache budget from $B=32$ to $B=4$.
This lowers the cache hit rate from 85.3\% to 53.7\%, 
}
\begin{wraptable}{r}{0.6\linewidth}
    \centering
    \vspace{-8pt}
    \caption{\rebuttal{Drafting on the host CPU for Gemma-4-26B-A4B-it (greedy, $\gamma=16$, 2048 new tokens, compiled HF Transformers). Averaged over the same benchmarks as in Table~\ref{tab:main}.}}
    \label{tab:cpu_drafter_main}
    \setlength{\tabcolsep}{3pt}
    \resizebox{\linewidth}{!}{%
    \begin{tabular}{l rrrr}
    \toprule
    \bf Method & \bf Hit (\%) & $\tau$ & \bf Speedup & \bf Gain \\
    \midrule
    D-PACE                         & --   & 7.00 & 3.39 & -- \\
    \name{} ($B=32$, 2 GPUs)       & 85.3 & 7.20 & 3.81 & \textcolor{green!50!black}{+12.5\%} \\
    \name{} ($B=4$, CPU drafter)   & 53.7 & 7.08 & 3.49 & \textcolor{green!50!black}{+3.1\%} \\
    \bottomrule
    \end{tabular}%
    }
    \vspace{-5pt}
\end{wraptable}
\rebuttal{
and with it the acceptance-length gains of looped drafting.
We thus fall back to a standard D-PACE drafter on cache misses.
Table~\ref{tab:cpu_drafter_main} shows that this CPU deployment still improves the speedup over the D-PACE baseline by 3.1\%, without an additional accelerator.
See Appendix~\ref{appendix:cpu} for detailed setup and results.
}

\subsection{Ablations}
We ablate our method on Qwen3-8B, greedy decoding, on MATH500, HumanEval, and MT-Bench. 

\begin{wrapfigure}{r}{0.5\linewidth}
    \centering
    \vspace{-10pt}
    \includegraphics[width=\linewidth]{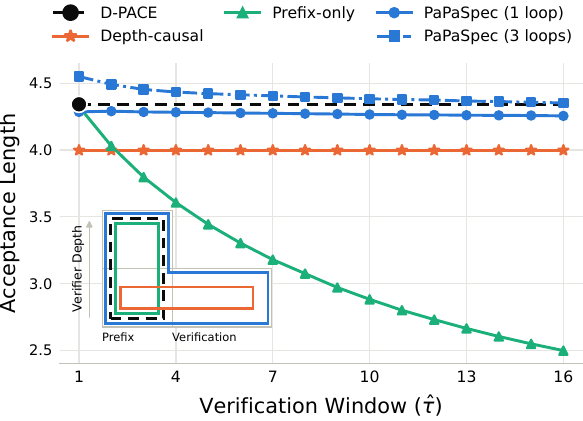}
    \vspace{-10pt}
    \caption{\rebuttal{Target conditioning ablation.}}
    \label{fig:acceptance-by-window}
\end{wrapfigure}
Figure~\ref{fig:acceptance-by-window} shows how different conditionings affect the average acceptance length as the speculated accepted position of the previous verification ($\hat\tau$) grows.
The drafters shown here are all trained on the 100k subset.
DFlash/D-PACE's rich conditioning on all verifier hidden states yields the highest acceptance length, but requires the verifier's forward pass to complete first and therefore only supports macro-sequential decoding.
Conditioning only on the prefix, i.e., the latents available before the current verification round began (prefix-only), supports macro-parallel decoding but degrades sharply as $\hat\tau$ grows.
Conditioning only on the layers that have so far been verified (depth-causal, i.e., causally restricting conditioning across both prefix and verification window) increases the acceptance length. 
\name with full hybrid conditioning further reduces the gap to D-PACE, which can be closed with looping, while supporting macro-parallel decoding.\looseness-1

\paragraph{Looping.} 
\begin{wrapfigure}{r}{0.5\linewidth}
    \centering
    \vspace{-10pt}
    \includegraphics[width=\linewidth]{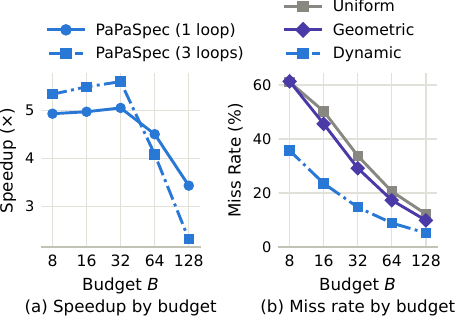}
    \caption{\rebuttal{Budget allocation ablation.}}
    \label{fig:budget-curves}
\end{wrapfigure}
We have already demonstrated looping's positive impact on the average acceptance length.
Figure~\ref{fig:budget-curves}a shows that the benefit transfers to actual speedup improvements.
We analyze the speedup as a function of the cache budget $B$.
Increasing the budget improves the speedup as it reduces cache misses, but benefits degrade at large budgets that push the drafting into a compute-bound regime.
This transition happens earlier for looped drafters due to their increased per-token compute. 
Despite saturating earlier, the peak speedup of looped drafters still exceeds that of single-loop drafters.
\rebuttal{Additionally, scaling the drafter's compute through looping rather than a deeper architecture provides better speedups (Appendix~\ref{appendix:loopvsdeep}).}
Overall, this result shows that \emph{investing compute in looping is more beneficial than naïvely increasing the draft budget with a single loop.}\looseness-1

\paragraph{SSD cache.} 
Figure \ref{fig:budget-curves}b further compares our dynamic confidence-based cache allocation with the uniform and geometric baselines. 
Our method reduces the miss rate for all cache budgets, with particularly pronounced gains in low budget regime: 
For example, at a cache budget of 8, our method reduces the miss rate from 61.2\% to 35.8\%. 
By extending the cache budget from 8 to 32, we can further reduce the miss rate to 14.7\%. 
More detailed results are in Appendix~\ref{appendix:additional-main-results}.\looseness-1

\section{Related work}

\paragraph{Speculative drafters} 
Although multi-token prediction (MTP)~\citep{stern_blockwise_2018, gloeckle_better_2024} already used the target's final-layer hidden state, initial speculative decoding used disconnected target and draft models~\citep{leviathan_fast_2023,chen_accelerating_2023}.
To avoid the need for a separate drafter, self-speculative decoding either adopts sub-networks of the target model~\citep{liu_kangaroo_2024, zhang_draft_2024}, turns the target model itself into a parallel sampler~\citep{monea_pass_2023},
or generates candidate continuations in parallel via Jacobi forcing~\citep{fu_break_2024}.
Despite requiring little or no adaptation, 
these methods yield limited speedups. 
Hence, efforts turned to designing tailored drafters with variable levels of target integration.
While some approaches maintained loose conditioning~\citep{an_pard_2026, an_pard2_2026}, 
major efforts went into drafters conditioned on the target's final hidden state, including Medusa~\citep{cai_medusa_2024}, Hydra~\citep{ankner_hydra_2024}, ParallelSpec~\citep{xiao_parallelspec_2024}, Falcon~\citep{gao_falcon_2025}, EAGLE~\citep{li_eagle_2024}, and EAGLE-2~\citep{li_eagle2_2024}. 
EAGLE-3~\citep{li_eagle3_2025} tightened conditioning by accessing target latents from multiple layers, achieving significantly higher acceptance rates. 
SpecDiff~\citep{christopher_speculative_2025a} introduced discrete diffusion to drafters, while SpecDiff-2~\citep{sandler_specdiff2_2026} improved their alignment to the target with new loss functions; however, the diffusion drafters are fully decoupled. 
P-EAGLE~\citep{hui_peagle_2026} parallelizes target-latent-conditioned AR drafting, using block diffusion modeling
with target-latent conditioning by superposing token embeddings with conditioning states at the already decoded token positions. 
DFlash~\citep{chen_dflash_2026} uses block diffusion modeling 
with multi-layer target conditioning. 
DFlare~\citep{zhang_dflare_2026} increases the flexibility by generating drafter-layer-specific conditionings.
All prior latent-conditioned drafters require the target's full latent computation before drafting can begin, which \name resolves with prefix and depth-causal conditioning. 
\looseness-1

\paragraph{Macro-parallel verification and drafting}
AMUSD~\citep{mcdanel_amusd_2025} asynchronously performs drafting and verification on separate devices, rolling back the draft from the point of disagreement whenever the verifier rejects it.
SSD~\citep{kumar_speculative_2026} drafts possible alternative outcomes for each position, allocating a budget based on a geometric distribution over acceptance length.
Motivated by different GPU count scaling behavior of target and draft models, SwiftSpec~\citep{zhang_swiftspec_2026} distributes them across separate GPUs. 
Mirror-SD~\citep{bhendawade_mirror_2025} speculates the next prefix by reading out intermediate verifier states; however, the subsequent drafting itself is performed with a micro-sequential drafter and without any target conditioning. 
All these macro-parallel drafters remain micro-sequential and lack target-latent conditioning.
Yet, tight conditioning is known to yield the highest acceptance length, \rebuttal{as we show in Appendix~\ref{appendix:acceptlength}.}

\textbf{Looping} 
Introduced by~\citet{dehghani_universal_2019}, looping has shown benefits in inference-compute scaling at constant parameter count~\citep{geiping_scaling_2025,wang_hierarchical_2025,jolicoeur-martineau_less_2025, zhu_scaling_2026, prairie_parcae_2026, mcleish_teaching_2026}. 
The success sparked interest in accelerating looped architectures themselves, as shown in contemporary works on pipelined self-speculative decoding~\citep{cho_loopspec_2026, ha_wavefront_2026}. 
However, these works can only accelerate looped Transformers. 
Recently, looping has been proposed as a depth-scaling mechanism in diffusion models~\citep{lee_looped_2026, carballo-castro_scaling_2026}.
However, none of these works considered looping to improve the acceptance length in speculative drafters. 

\section{Conclusion}
We presented \name, a speculative decoding approach featuring a target-latent-conditioned parallel drafter that operates concurrently with the target verification. 
To achieve this, our depth-causal conditioning allows the drafter to consume the target latents that have been computed so far during verification. 
Thanks to removing drafting from the critical latency path, the drafter can dynamically increase its compute budget via looping. 
When integrated into SSD and enhanced with a novel dynamic drafting-budget allocation, \name achieves up to \rebuttal{31}\% higher speedups than DFlash.\looseness-1 

\paragraph{Limitations and future work}
In our experimental setup, we follow the established practice of operating the drafter on an additional GPU of the same size as the target host~\citep{kumar_speculative_2026}.
With the planned future general availability of accelerators for drafting~\citep{bartlett_corsair_2026, weidman_groq3lpx_2026}, future work could demonstrate \name's performance at lower hardware cost. 
\rebuttal{
Preliminary indications from drafting on the CPU already show promise in this direction. 
}

\subsection*{AI use statement}
In this work, we used generative AI tools to implement parts of the software code and suggestions in experimental design and feedback on our research methodology.
We have not used generative AI tools for data analysis and interpreting results,
and its use in assistance in proofs or mathematical claims are not applicable to this work.
Additionally, we used generative AI tools for formatting and discovering existing literature. We have reviewed all AI-assisted work by manually verifying all artifacts.
We take responsibility for the final content of this work,
including text, claims or artifacts produced with the aid of generative AI.

\subsection*{Ethics statement}
Our method accelerates LLM inference while provably preserving the output distribution of the target model, a property we inherit from speculative decoding. It changes only how fast tokens are generated, not which tokens are generated. It therefore introduces no new capabilities, biases, or risks, and the ethical concerns associated with LLMs (e.g., harmful content, bias, privacy) remain properties of the target model and fall outside the scope of this work. Our experiments use only publicly available models and datasets in accordance with their licenses.

\subsection*{Reproducibility statement}
All training datasets, benchmarks, and models are public. We describe our experimental setup and implementations in Appendix \ref{appendix:experimental-details}. We report the computational resources required in \ref{appendix:compute} to reproduce the training of all drafters. We then describe specifics on training details in Appendix \ref{appendix:training}. We summarize the architectural parameters and statistics of all the models used in this work in Appendix \ref{appendix:hyperparams}. For key concepts of \name{}, we provide code snippets in Appendix \ref{appendix:impl-details}.
As indicated by ICLR's author guidelines, we are available to share the code anonymously during the rebuttal period.


\bibliography{misc, hd-learning}
\bibliographystyle{iclr2027_conference}

\newpage
\appendix
\section{\rebuttal{Extended main results}}

\subsection{\rebuttal{Acceptance length and SSD baseline}}\label{appendix:acceptlength}
\input{tables/main_table_extended}
\rebuttal{\paragraph{Acceptance lengths.} Table~\ref{tab:main_tau} additionally includes the acceptance length of our main results, alongside the speedups. Our model achieves a higher acceptance length than our strong baseline of D-PACE, despite the depth-causal conditioning. The increase in acceptance length is one of the two factors responsible for \name's speedups over DFlash and D-PACE, the other being hiding the drafting latency. 
The key enabler to \name's higher acceptance length is looping, as ablated in Figure~\ref{fig:acceptance-by-window} which showed that \name without looping results in an acceptance length slightly below D-PACE. On the other hand, \name with looping boosts the acceptance length above D-PACE, especially when the block under verification achieved a small acceptance length and therefore more hidden states are available of the most recent context.}

\rebuttal{\paragraph{SSD baselines.} For a comparison with a macro-parallel method, we chose to compare against Speculative Speculative Decoding (SSD) ~\citep{kumar_speculative_2026}.
We reimplemented SSD in our own HF Transformers inference framework so that all SSD, DFlash, D-PACE, and \name share the same verification code, AR baseline, and timing measurements. Like the original SSD work, the drafter runs on a separate GPU while the target verifies.
We restricted our evaluation to Qwen3-8B as the target model and benchmarked two smaller drafters from the same family: Qwen3-0.6B and Qwen3-1.7B.
In order to have the strongest possible baseline of SSD in our setting, each of the two drafters was tuned separately using a subset of 32 samples from each of MATH500, HumanEval, and MT-Bench. SSD sets the number of tokens $K$ drafted per step, equivalent to $\gamma$ in block drafting, such that drafting finishes within one verification step. In the original setting, it was set to $K=5$, but because our smaller target verifies much faster, we swept $K \in \{1, 2, 3, 4, 5, 6, 8\}$.
For each K, we also varied the cache budget. We either tried the same number (1, 2, or 3) at every position, the uniform allocation strategy, or the geometric allocation strategy proposed in the SSD paper at comparable sizes.
At temperature 1, we also tested SSD's Saguaro sampling with $C \in \{0.3, 0.6\}$ against no adjustment. For each drafter and temperature, we kept the configuration with the highest mean speedup over the three tuning datasets. Table~\ref{tab:ssd-selected} shows the selected parameters for the baseline. $K=2$ was selected in all four cases, with a budget of 6 (for temperature 0, Qwen3-1.7B) to 9 (for all 3 others), and Saguaro sampling ($C=0.3$) was selected only for the 1.7B drafter at temperature 1. 
\begin{table}[t]
\centering
\caption{\rebuttal{SSD configurations selected by the tuning sweep (highest mean speedup over MATH-500, HumanEval and MT-Bench).}}
\label{tab:ssd-selected}
\begin{tabular}{llcccc}
\toprule
Drafter & Temp. & $K$ & Strategy & Cache Budget & Saguaro $C$ \\
\midrule
Qwen3-0.6B & 0 & 2 & Uniform   & 9  & --  \\
           & 1 & 2 & Geometric & 9  & off \\
\midrule
Qwen3-1.7B & 0 & 2 & Uniform   & 6  & --  \\
           & 1 & 2 & Uniform   & 9  & 0.3 \\
\bottomrule
\end{tabular}
\end{table}
The selected configurations were then evaluated as all other methods in our main table. Seven benchmarks, 128 prompts each, up to 2048 new tokens, at temperatures 0 and 1, with all speedups measured against an autoregressive baseline on 8 H100s.
}

\subsection{Throughput comparison to vLLM}
\label{appendix:vllm}

\input{tables/servings_hf_vllm}

Table~\ref{tab:serving_stacks} \rebuttal{compares the decoding speeds between the two inference backends used in this work}. The first is \rebuttal{vLLM (v0.31)} and the second is HuggingFace Transformers under \texttt{torch.compile}, as used in the main experiments and ablations in this work.
We benchmark the same checkpoints, for 128 prompts per dataset, across Math500, HumanEval, and MT-Bench, with a maximum of 2048 new tokens, on H100 GPUs. 
We use separate dynamic budgets for the two backends, as we found that 32 provides the best speedup for \name served with HF Transformers while 16 is best for vLLM serving. 

\rebuttal{Comparing the decoding speeds of HF Transformers with vLLM on Qwen3-8B, we find that indeed HF Transformers is competitive at 718 tok/s but falls 16\% short of vLLM's 858 tok/s.}

\rebuttal{The speedup gain of \name over DFlash differs between the two backends, with the HF backend serving attaining higher gains over vLLM. One reason for this is that the optimized vLLM serving reduces the memory-bandwidth bottleneck and shifts the compute-bound regime earlier than the HF backend. This means that the drafter has less time to complete before its latency is no longer fully hidden off the critical path. This is also why we found that vLLM reached higher speedups at the smaller dynamic cache budget of 16, which meant a shorter acceptance length in vLLM servings compared to HF Transformers.}

\newpage
\subsection{\rebuttal{Concurrent serving}}\label{appendix:batched}

\rebuttal{
This appendix extends \name{} from single-request decoding to concurrent serving of $c > 1$ requests.
In the single-request setting, \name{} provides a two-fold gain over macro-sequential drafters: the drafting latency is removed from the critical path, and the drafting time gained is spent on looping to increase the acceptance length.
The latency gain, however, fully relies on a cache hit.
Under concurrent serving, the target model verifies all $c$ requests in a single batched forward pass, so the next verification can only be issued once a draft is available for every request in the batch.
A single cache miss therefore forces a macro-sequential drafting step for the entire batch.
Assuming independent misses with a per-request miss rate $p_\text{miss}$, the probability that a decoding step proceeds fully macro-parallel is $\Pr(\text{batch hit}) = (1 - p_\text{miss})^{c}$,
which decays exponentially with the concurrency $c$.
For example, even at a relatively low miss rate of 15\%, only 27\% of the decoding steps would be macro-parallel for $c=8$, and fewer than 8\% for $c=16$.
Macro-sequential drafting would hence be required in almost every decoding step, and \name's latency gain vanishes.
As a remedy, we extend \name to draft without an anchor token, which removes cache misses altogether.
}

\paragraph{Architecture}
\rebuttal{
Recall that for a speculated acceptance length $\hat\tau$ and a speculated bonus (anchor) token $b$, the drafter input is
$z = \big(\mathrm{E}(b), \mathrm{E}([M]), \dots, \mathrm{E}([M])\big) \in \R^{(\gamma+1)\times d}$.
We augment \name{} to draft not only from speculated anchor tokens, but also from a generic placeholder token $[P]$ with a trainable embedding $e_{[P]} \in \R^{d}$, initialized as the embedding of the mask token:
\begin{equation*}
  z_{[P]} = \big(e_{[P]}, \mathrm{E}([M]), \dots, \mathrm{E}([M])\big).
\end{equation*}
Since $e_{[P]}$ is independent of the position and content, a single placeholder can be reused for every speculated acceptance length $\hat\tau$.
Replacing the anchor token, however, removes valuable information for drafting.
\name{} instead has to infer the content of the anchor from the target latents, which become increasingly available as verification progresses through the prefix and depth-causal conditioning.
Notably, the bonus token itself is only determined at the end of the verification phase.
Aiming to recover some information that has been lost, we introduce a small correction head~\citep{cheng_dspark_2026}, which edits the drafter's marginals autoregressively, starting from the actual bonus token.
For brevity, we index positions within the drafted block, with $x_0 \in \{b, [P]\}$ denoting the anchor and $x_{1:\gamma}$ the drafted tokens, and we omit the conditioning on the prefix and target latents.
Let $q_i(\cdot \mid [P])$ be the drafter's marginal at position $i$, and $x_0 = b$ the bonus token returned by verification.
The corrected block distribution factorizes as
\begin{align*}
  \tilde q\big(x_{1:\gamma} \mid [P], x_0\big) &= \prod_{i=1}^{\gamma} \tilde q_i\big(x_i \mid x_{i-1}, [P]\big),
  \\
  \tilde q_i\big(v \mid x_{i-1}, [P]\big) &=
  \frac{\exp\!\big(\log q_i(v \mid [P]) + g(x_{i-1}, v)\big)}
       {\sum_{v' \in \mathcal{V}} \exp\!\big(\log q_i(v' \mid [P]) + g(x_{i-1}, v')\big)},
\end{align*}
where $g(u, v) = \big(U_\text{in}[u]\, U_\text{out}\big)_v$ is a low-rank bigram bias with $U_\text{in} \in \R^{|\mathcal{V}| \times r}$, $U_\text{out} \in \R^{r \times |\mathcal{V}|}$, and $U_\text{in}[u]$ the row of token $u$.
Note that the chain always starts from the actual bonus token $x_0 = b$: the head is what brings $b$ into the draft.
At inference, the draft is decoded greedily along the chain, $x_i = \arg\max_v \tilde q_i(v \mid x_{i-1}, [P])$.
Since the marginals $q_i$ are computed in a single parallel forward pass before $b$ is known, the correction adds only $\gamma$ sequential steps after verification, each a lookup of $U_\text{in}[x_{i-1}]$, a rank-$r$ projection to the vocabulary and an argmax.
We find that a rank of $r=64$ is sufficient.
}

\paragraph{Unified cache}
\rebuttal{
We place one placeholder block at every possible acceptance length.
The placeholder cache can be extended with drafts from actual bonus tokens, e.g., the top-$B'$ entries of our confidence-based budget allocation. 
Drafts from actual bonus tokens are preferred whenever available, and every outcome is covered, i.e., $p_\text{miss} = 0$ and $\Pr(\text{batch hit}) = 1$ for any $c$.
}

\paragraph{Training}
\rebuttal{
We initialize from a pre-trained vanilla \name{} drafter and fine-tune it jointly with the placeholder embedding and the optional Markov head for two epochs on the same data.
As in vanilla \name{}, 20\% of the optimizer steps use exact anchors without looping, which preserves the macro-sequential drafting needed for a request's first block.
The remaining steps use $R=3$ loops. 
With probability $\rho$, the block is then trained as the placeholder block; otherwise it keeps its exact anchor, which trains the looped exact-anchor entries of the unified cache.
$\rho$ ramps linearly from $0$ to $0.8$ over the first 10\% of the training steps.
The D-PACE loss is applied to the corrected distribution $\tilde q$, teacher-forced on the ground-truth tokens; $U_\text{out}$ is zero-initialized so that the Markov head starts as the identity.
The backbone uses a peak learning rate of $1.5\times10^{-4}$, and the new parameters ($e_{[P]}$, $U_\text{in}$, $U_\text{out}$) use $6\times10^{-4}$, both with 3\% linear warmup followed by cosine decay.
}

\paragraph{Inference}
\rebuttal{
We integrate \name{}'s concurrent serving into vLLM.
A new request has no cache yet, so its first block is drafted macro-sequentially from the exact anchor without looping.
Afterwards, the placeholder cache alone requires $\gamma+1$ drafts per request, so the drafter processes $c\cdot(\gamma+1+B')$ blocks per verification step, which grows linearly with the concurrency $c$.
We use two measures to keep drafting hidden behind verification.
First, we reduce the number of additional exact-anchor drafts $B'$ in the unified cache as $c$ grows.
Second, we optionally defer the LM head until after verification.
During verification, the drafter computes the final latents of all cached blocks.
Once the outcome $(\tau, b)$ is known, the LM head is applied only to the matching block, reducing the LM-head workload from $c\cdot(\gamma+1+B')$ to $c$ blocks per step.
At the highest concurrency, we additionally defer the drafter's last layer, which is then likewise applied to the matching block only.
For example, on Gemma-4-26B-A4B we use $B'=6$ up to $c=16$, $B'=2$ at $c=32$, and $B'=0$ at $c=64$; the LM head is deferred from $c\geq16$ and the last layer at $c=64$.
The Markov head, which conditions on the newest bonus token $b$, is likewise applied after verification and therefore always lies on the critical path; due to its low rank $r$, its latency overhead remains small.
}

\input{tables/serving_concurrent}

\paragraph{Results}
\rebuttal{
We evaluate on the throughput splits of SPEED-Bench~\citep{abramovich_speedbench_2026} with 1k and 2k input tokens, which test realistic serving scenarios across diverse semantic regimes.
As we can see, against our D-PACE baseline trained on the same data, our \name{} achieves the highest throughput up to a concurrency of 16.
Removing the Markov head still yields a gain over D-PACE at lower concurrencies; however, its acceptance length is lower, which the saved synchronous drafting latency cannot compensate for at higher concurrencies.
One interesting observation is that \name{} is above D-PACE again at $c=64$. We attribute this to Gemma's increased verification latency under concurrent MoE serving, which buys \name{} additional drafting time.
Finally, we observe that \name{}'s gains grow with longer input lengths at low to medium concurrency; this may be due to the increased macro-sequential compute of D-PACE, which our \name{} can hide.
}

\subsection{\rebuttal{Asymmetric compute resources: Drafting on CPU}}\label{appendix:cpu}
\rebuttal{
Our main \name experiments utilize an additional equally sized H100 GPU for macro-parallel drafting.
Given that all the considered target models fit on a single H100 GPU, verification and drafting compute is symmetric.
Here, we showcase that drafting compute requirements can be significantly lower by performing the macro-parallel drafting on the CPU cores provided by the same compute node.
}
\paragraph{Setup}
\rebuttal{
We perform the CPU deployment study on Gemma-4-26B-A4B-it, the largest target model we consider, which nonetheless pairs with a small drafter.
We use the node's two Intel Xeon Platinum 8468 CPUs, which provide 96 cores across two sockets.
The drafter runs tensor-parallel over the two sockets, with one process and 45 threads per socket; two cores are reserved for the process driving the verifier GPU.
For efficient memory management and tiling, the drafter runs in a tailored C++ implementation on the CPUs' AMX matrix units.
Linear layers (attention projections and FFNs) are computed in W8A8 precision, while attention remains in BF16 with FP32 accumulation.
To further reduce the CPU workload, we offload parts of the drafter to the verifier GPU: the computation of the conditioning signals $c^{(l)}_t$, the last drafter layer $\mathcal{D}^{(L_D)}$, and the LM head.
The last drafter layer is the only one that depends on the final conditioning signal, which becomes available only at the end of verification; the CPU therefore stops before it and pushes the residual stream of all $B$ branches to the GPU through shared memory.
There, the last drafter layer runs over all branches concurrently with the final target layers, and the LM head processes the verification block and the branches' drafts jointly in a single batched call, so its weights are loaded only once.
On a cache hit, the next block is read out from its branch's rows without any further GPU work on the critical path.
Running the last drafter layer and the additional LM head rows concurrently with verification increases the verification latency by only 0.15\,ms, about 1.4\% of the 11\,ms verification time of Gemma-4-26B-A4B-it.
The macro-parallel drafts run the drafter's looped layers with three iterations.
We use a fixed cache budget of $B=4$ branches.
Such a small budget lowers the cache hit rate, and with it both the latency gain and the acceptance length.
We compensate by using a standard D-PACE drafter for the synchronous drafts after a cache miss.
Since these drafts are computed on the verifier GPU anyway, a dedicated low-latency drafter with a high acceptance length is the natural choice for them.
}

\paragraph{Results}
\input{tables/cpu}
\rebuttal{
Table~\ref{tab:cpu_drafter} shows the detailed results.
To rule out variability between nodes, D-PACE and the CPU version of \name{} are measured in the same run on the same node.
Using only a single GPU and offloading the macro-parallel drafting to the host CPU, \name{} improves the speedup over D-PACE by 3.1\% on average, consistently across all seven benchmarks.
Compared to drafting on a second GPU (+12.5\%), the smaller gain stems from the reduced cache budget ($B=4$ instead of $32$), which lowers the cache hit rate from 85.3\% to 53.7\%.
On a cache miss, \name{} falls back to synchronous D-PACE drafting, so the looped macro-parallel drafts with higher accuracy are used less often.
This is reflected in a lower acceptance length (7.08 vs.\ 7.20), which nonetheless remains above that of D-PACE (7.00).
Yet, this preliminary implementation shows that spending otherwise idle compute on macro-parallel drafting is beneficial, even when that compute comes only from the host CPU.
}

\section{Additional Results and Ablations}

\subsection{Latency breakdown of \name{}}
\label{appendix:ssd-timeline-appendix}

\begin{figure*}\centering
\begin{minipage}[t]{0.39\linewidth}\centering
  \includegraphics[width=\linewidth]{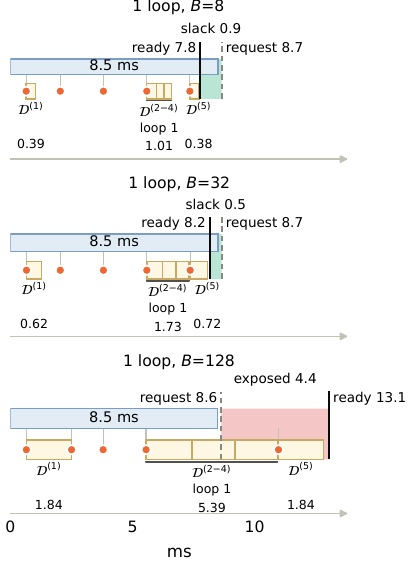}\\[2pt]{\small (a) 1 loop}
\end{minipage}%
\begin{minipage}[t]{0.61\linewidth}\centering
  \includegraphics[width=\linewidth]{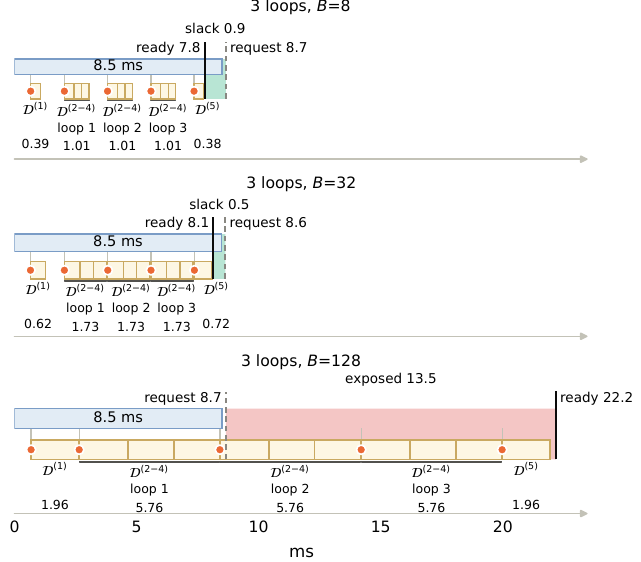}\\[2pt]{\small (b) 3 loops}
\end{minipage}
\caption{\rebuttal{Event timeline across different cache budgets for the Qwen3-8B target, operating our drafter with \textbf{(a)} 1 loop and \textbf{(b)} 3 loops. The green area indicates the slack where the draft is ready before the target model requests the next draft tokens. Red indicates the drafting latency is exposed on the critical path, with the draft being ready only after it is requested by the target.}}
\label{fig:timeline-appendix}
\end{figure*}

Figure \ref{fig:timeline-appendix} covers the latencies of key events in a single step of parallel verification and drafting in \name{} across cache allocation budgets. The latency events are invariant to the dataset being benchmarked; they depend only on the drafter, the target model, and key parameters to our method such as the number of loops and budget. For \name{} with a single loop, the drafter prepares all draft tokens 0.9ms before the target requests the next batch of draft tokens for a budget of 8, and 0.5ms for a budget of 32. However, with a budget of 128, the compute is saturated, and the drafts are ready 13.1ms after the target requests them. This is especially pronounced for \name{} with 3 loops, where the cache is only ready 13.5ms after being requested by the verifier, more than the verification time itself. Hence, the cache budget and number of loops have to be jointly considered given the compute resources. 

\subsection{Budget allocation}\label{appendix:additional-main-results}
\input{tables/ablation_cache_loop}

\paragraph{Cache strategy ablation} The detailed results of our ablation of the three drafter cache allocation strategies are in Table~\ref{tab:ablation-cache-loop}. The dynamic cache strategy consistently outperforms all other cache strategies in miss rate and speedup. 
Looping further improves speedup through two factors: increasing the acceptance length and decreasing the miss rate. The speedup eventually saturates after a budget of 32 for \name{} with 3 loops and after a budget of 64 for 1 loop, mainly due to the drafter taking more time than the verifier to prepare the drafts, exposing drafting along the critical path.
\begin{figure}[t]
    \centering
    \includegraphics[width=0.8\linewidth]{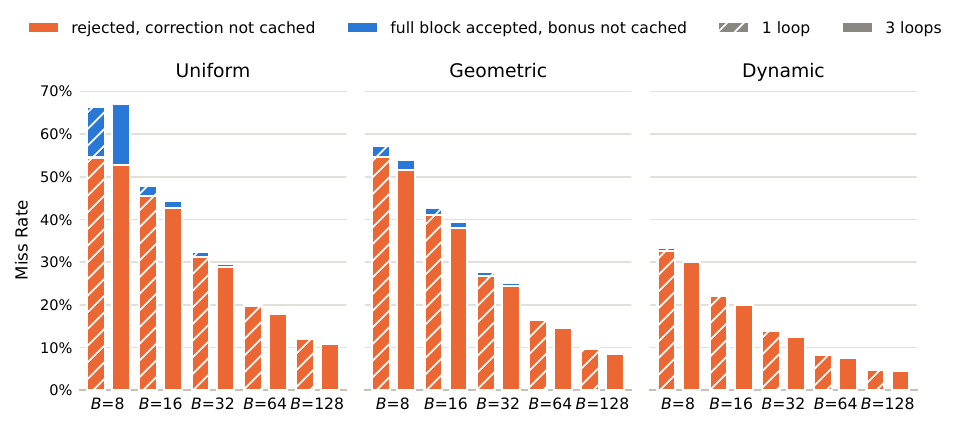}
    \caption{\rebuttal{Breakdown of cache miss due to rejection and full block acceptance.}  
    }
    \label{fig:ssd-loop-ablation}
\end{figure}
\paragraph{Miss Rate decomposition.} \rebuttal{Figure \ref{fig:ssd-loop-ablation} shows the decomposition of cache misses caused by the bonus tokens not cached at full block acceptances (blue) and blocks whose bonus token was not cached at positions earlier than $\gamma+1$. Interestingly, almost no cache misses are caused by not having cached the $\gamma+1$ token in full block acceptances with our dynamic strategy.
Looping, across the uniform and geometric cache strategies, slightly reduces the cache miss rate.}

\subsection{\rebuttal{Looping vs Depth ablation}}
\label{appendix:loopvsdeep}

\rebuttal{\name{} enables the latency cost of drafting to sit fully behind verification. 
As such, an alternative scaling mechanism to looping is to increase the parameter count of the drafter through a deeper architecture. 
We ablated this by training deeper drafters and comparing against \name with 1 to 3 loops. All models are trained on our ablation training dataset consisting of 100k samples and benchmarked with the HF Transformers backend.
}

\rebuttal{In Table~\ref{tab:depth_vs_loop}, we show the speedup and acceptance lengths when serving the drafters macro-sequentially on a single H100 and macro-parallel using a second H100 to build the drafter's dynamic cache.
In macro-sequential drafting, where drafting latency sits on the critical path, we find the highest speedup to be attributed to the fastest drafter with only 5 forward passes. Additional forward passes, whether through looping or a deeper network, do not bring an overall speedup.}

\rebuttal{For macro-parallel drafting however, both loops and deeper networks do slightly increase the speedup. More importantly, when comparing \name with looping its 3 middle layers, we outperform the deeper drafter counterpart at iso-forward passes during drafting. As used in the main results of this work, the highest speedup comes from \name with 5 decoder layers and 3 loops, with a total of 11 decoder layer forward passes at a speedup of 4.36$\times$. The deeper 11-layer decoder drafter trails at 4.26x speedup. This evidence motivates our choice of extending the newly available drafter compute with looping rather than a deeper drafter.}

\input{tables/loopvsdeep}

\subsection{\rebuttal{Mixing weights}}
\rebuttal{
We ablate the mixing weights in the conditioning vector $c^{(l)}_{t}$ by comparing three parameterizations in Table~\ref{tab:mixing_weights}.
In the per-layer, each of the $m$ drafter layers learns its own weights $\alpha^{(l)}, \beta^{(l)} \in \mathbb{R}^{m}$, which amounts to $2m^{2} = 50$ mixing weights for the setting in this paper of $m = 5$.
In the layer-tied variant, a single pair $\alpha, \beta \in \mathbb{R}^{m}$ is used by all drafter layers which amounts to $2m = 10$ mixing weights.
In the third variant, we remove the mixing weights entirely so that $c^{(l)}_{t}$ reduces to an unweighted sum of the projected latents $W^{(i)} h_{t}^{(l_i)}$.}

\rebuttal{We trained three drafters for each of these three conditioning methods on the 100k sample training dataset.
Removing the mixing weights entirely reduces the mean acceptance length over MATH-500, HumanEval and MT-Bench, from 5.20 for the per-layer variant and 5.23 for the layer-tied variant to 5.18, which indicates that mixing weights are useful in slightly boosting acceptance length.
The layer-tied variant attains a slightly higher acceptance length than the per-layer variant with one fifth of the mixing weights.
}

\begin{table}[t]
    \centering
    \small
    \caption{
    \rebuttal{Ablation of the mixing weights $\alpha, \beta$ in the definition of $c^{(l)}_{t}$. Mean acceptance length of the Qwen3-8B drafter (block size 16, greedy decoding), averaged over MATH-500, HumanEval, and MT-Bench.}
    }
    \begin{tabular}{lcc}
        \toprule
        Mixing weights & \# Mixing weights & Acceptance length \\
        \midrule
        None ($\alpha_{i} = \beta_{i} = 1$)                                & 0  & 5.18 \\
        Per-layer $\alpha^{(l)}, \beta^{(l)}$ & 50 & 5.20 \\
        Layer-tied $\alpha, \beta$                                             & 10 & \textbf{5.23} \\
        \bottomrule
    \end{tabular}
    \label{tab:mixing_weights}
\end{table}

\section{Experimental Details}
\label{appendix:experimental-details}

\subsection{Compute Resources}
\label{appendix:compute}
We used a full node of 8xH100 80GB HBM3 GPUs for all our training runs. The training of our ablations on the 100k dataset, all of which we were using Qwen3-8B as the target model, took 6.5 hours, for a total of 52 GPU hours. As for our \name{} models reported in the main results, the drafter for Gemma-4-26B-A4B took 66 hours to train (528 GPU hours), the drafter for Qwen3-4B took 36 hours (288 GPU hours), and the drafter for Qwen3-8B took 61 hours (488 GPU hours). 

\subsection{Training Implementation}
\label{appendix:training}
\paragraph{Framework} We adapted the official implementation of DFlash's training recipe from vLLM's Speculators \citep{speculators2025}, for both the reconstruction of the D-PACE baseline, reproducing the released checkpoint, and the \name{} checkpoints. We cap the sequence length of each prompt in our data mixture to 3072. We sample 512 anchors per sequence and use a block size $\gamma$ of 16.
\paragraph{Optimizer and batch size} We train up to 6 epochs and use AdamW with a learning rate of 6e-4, a linear warmup of 4\% of steps, a clipping threshold of 1 and a cosine scheduler. 
We use a global batch size of 8 sequences per optimizer step, one sequence per 8 GPUs and aggregate with Distributed Data Parallel.
To reproduce the reported acceptance length metrics reported in DFlash \citep{chen_dflash_2026}, we additionally needed to cast the FP16 weights up to FP32 for AdamW, with the forward pass and the checkpoint being kept in FP16. This was borrowed from SpecForge's implementation \citep{li2026specforge}.
\paragraph{Training dataset} For the ablation experiments, in Figure \ref{fig:acceptance-by-window}, we use a mixture of 50k prompts from CodeAlpaca~\citep{luo2024wizardcoder} and 50k prompts from NVIDIA's Nemotron Post-Training Dataset V2~\citep{NemotronPostTrainingDatasetV2}. All 50k samples from CodeAlpaca are code-related, whereas for the Nemotron Post-Training Dataset we sample with equal probability from their chat, STEM, math, and code. We withheld 10\% of the training data for validation metrics during training. 
%
Our main \name{} results, as well as our D-PACE baseline and a reproduction of DFlash matching the reported acceptance lengths, were trained on all of CodeAlpaca (111k rows) and 800k samples from the Nemotron Post-Training dataset. More specifically, 200k samples in STEM, 200k in math, 175k in code, and 225k in chat. 
To obtain drafters tailored to each target model, we generate on-policy ground-truth responses with each target model. 
\paragraph{Hidden states} Generating the verifier hidden states for training can be done either offline, where all hidden states are generated and stored for later use during training, or generated online when needed during training. The total amount of hidden states depends multiplicatively on the sequence length, number of samples in the training dataset, the verifier hidden state dimension and the amount of hidden states $m$ we want to condition on. In our experiments, we have a maximum sequence length of 3072, 820\,144 training sequences, and a hidden dimension of 4\,096 for Qwen3-8B, 5 target hidden layers and FP16 precision (2 bytes per value) it amounts to $3072 \cdot 820,144 \cdot 4,096 \cdot 5 \cdot 2 \approx 103.2$ TB. This was prohibitively expensive to store, so we generated the hidden states online during training by launching a vLLM instance on each of our training GPUs to prefill and retrieve the hidden states. We allocated 30\% of each H100's 80 GB VRAM to vLLM extraction's and the rest for training. For Gemma-4-26B-A4B we had to host the vLLM servers on separate GPUs from training, allocating 4 GPUs to each, since both would not fit in the 80 GB of the H100's VRAM.

\paragraph{Depth-causal conditioning} During training, we sampled $\hat \tau$ from $\Pr(\hat \tau{=}k) = (1-\rho)\,\rho^{k-1}$ using a $\rho$ of 0.8 for the smaller scale ablations trained on the 100k dataset and with a $\rho$ of \rebuttal{0.9} for our main results on the 800k dataset. 
\paragraph{D-PACE loss} The D-PACE \citep{wu_dpace_2026} loss function is borrowed from the official implementation, set with a weight smoothing hyperparameter of 0.5. We use this for both the strong D-PACE baseline, as well as our \name{} drafters.

\subsection{Model Hyperparameters}
\label{appendix:hyperparams}
Table \ref{tab:architectural-params} summarizes key architectural differences across the three target models and drafters. The drafters are identical to the official DFlash releases\footnote{https://huggingface.co/z-lab/Qwen3-8B-DFlash-b16} \footnote{https://huggingface.co/z-lab/Qwen3-4B-DFlash-b16} \footnote{https://huggingface.co/z-lab/gemma-4-26B-A4B-it-DFlash} for the corresponding target models. Our \name{} inherits the same hyperparameters, with only the lightweight addition of the $\alpha$ and $\beta$ parameters for our depth-causal conditioning. Looping extends the test-time compute at inference time, but keeps the parameters fixed.
\input{tables/architectural_params}

\section{Implementation}
\label{appendix:impl-details}
We highlight more specific details of the implementation of two key parts of \name{}: the target-drafter communication through NCCL and the conditioning mechanism.

\subsection{Verifier-drafter communication}
\label{appendix:verifier-drafter-communication}
Listings~\ref{lst:ssd-verifier} and~\ref{lst:ssd-drafter} give the per-step protocol. The two ranks run on two GPUs of the same node, joined by a dedicated NCCL process group, and exchange only fixed-shape point-to-point messages. \texttt{link} denotes this group as seen from one rank with \texttt{send}/\texttt{recv} the blocking and \texttt{isend}/\texttt{irecv} the non-blocking point-to-point primitives of \texttt{torch.distributed} addressed to the peer rank.

Each step opens with a request from the verifier, the pair $(\hat{\tau}, x_{n+1})$ of acceptance length and bonus token, and a fixed-size reply from the drafter (a cache-hit flag and the drafted tokens $\hat{x}_{n+2:n+\gamma}$). 

The verification forward is captured as $m{+}1$ CUDA graphs cut at the selected layers $\{l_1, \dots, l_m\}$, so each latent $h^{(l_k)} \in \mathbb{R}^{\gamma \times d_T}$ is sent with a non-blocking \texttt{end} as soon as its segment has been replayed.

On the drafter rank, the receive of each latent is posted right before the drafter layer that consumes it, since NCCL matches point-to-point messages by issue order and \texttt{wait()} is a stream-level wait. $\mathcal{D}^{(k)}$ runs while the verifier is still computing $h^{(l_{k+1})}$. Receives are posted one at a time (a pending receive kernel contends with the drafter's compute), and the drafter never synchronizes its device, which would serialize it behind the verifier's forward. Once all $m$ latents have arrived, the drafter runs $\mathrm{LM}_{\text{head}}$ over its cached branches while polling for the next request, so a cache hit is answered with no GPU work on the critical path.

\input{listings/communication_logic}

\subsection{Conditioning Mechanism}
Listing~\ref{lst:ssd-conditioning} gives the prefix and depth-causal conditioning as computed. The $W^{(i)}$ are the column blocks of DFlash's fusion matrix $W$, so the projected latents are formed once per position and then mixed. 
\rebuttal{On top of DFlash, the drafter learns two weight vectors $a, b \in \mathbb{R}^{m}$ ($2m$ mixing weights in total), shared by all drafter decoder layers. 
Every layer $\mathcal{D}^{(l)}$ mixes the verified prefix with the same $\alpha = \mathrm{softmax}(a)$ over all $m$ verifier hidden states, and the $\hat{\tau}-1$ positions under verification with $\beta_{\cdot,l}$, the softmax of $b$ over $l_1, \dots, l_l$ only, where the entries for $i > l$ are masked to $-\infty$. 
Layers therefore differ only in how many latents they may read. Both vectors are zero-initialised (i.e.\ uniform), so $\alpha$ starts as DFlash's shared conditioning. 
At inference the prefix conditioning is cached across steps; the $\beta$-mixed rows are rebuilt for each block as its latents arrive.}

\input{listings/conditioning}

\end{document}

%% file: math_commands.tex
\usepackage{amsmath,amsfonts,bm}

\def\eqref#1{equation~\ref{#1}}

\def\1{\bm{1}}

\DeclareMathAlphabet{\mathsfit}{\encodingdefault}{\sfdefault}{m}{sl}
\SetMathAlphabet{\mathsfit}{bold}{\encodingdefault}{\sfdefault}{bx}{n}

\newcommand{\R}{\mathbb{R}}



%% file: tables/main_table_compiled.tex
\begin{table}[t]
\caption{Speedup over AR generation for different speculative decoding methods on H100 GPUs. All approaches are compiled into a single CUDA-graph and operate at block size $\gamma=16$ with a maximum sequence length of $2048$. \rebuttal{SSD, reimplemented in our framework, instead uses a standalone Qwen3-0.6B or Qwen3-1.7B drafter with lookahead $\gamma=2$.} \name{} uses a dynamic budget of 32 for all target models. $\dagger$ indicates a public checkpoint rather than training reproduction.}
\label{tab:main}
\begin{center}
\setlength{\tabcolsep}{3pt}
\resizebox{\textwidth}{!}{%
\begin{tabular}{ll r@{}l r@{}l r@{}l r@{}l r@{}l r@{}l r@{}l r@{}l c}
\toprule
\multirow{2}{*}{\bf Model} & \multirow{2}{*}{\bf Method}
  & \multicolumn{6}{c}{\textsc{Math}} & \multicolumn{6}{c}{\textsc{Code}} & \multicolumn{2}{c}{\textsc{Chat}} & \multicolumn{2}{c}{\multirow{2}{*}{\textit{Avg.}}} & \multirow{2}{*}{\begin{tabular}[c]{@{}c@{}}Gain of\\\name{}\end{tabular}} \\
\cmidrule(lr){3-8} \cmidrule(lr){9-14} \cmidrule(lr){15-16}
 & & \multicolumn{2}{c}{GSM8K} & \multicolumn{2}{c}{MATH-500} & \multicolumn{2}{c}{AIME26} & \multicolumn{2}{c}{HumanEval} & \multicolumn{2}{c}{MBPP} & \multicolumn{2}{c}{LCB} & \multicolumn{2}{c}{MT-Bench} & & & \\
\midrule
\multicolumn{19}{c}{Temperature = 0} \\
\midrule
\multirow{5}{*}{Qwen3-8B}
  & \rebuttal{SSD (0.6B)} & \rebuttal{2.45} & & \rebuttal{2.45} & & \rebuttal{2.36} & & \rebuttal{2.30} & & \rebuttal{2.25} & & \rebuttal{2.30} & & \rebuttal{1.87} & & \rebuttal{2.29} & & \rebuttal{\textcolor{green!50!black}{+161\%}} \\   
  & \rebuttal{SSD (1.7B)} & \rebuttal{2.00} & & \rebuttal{2.01} & & \rebuttal{1.94} & & \rebuttal{1.89} & & \rebuttal{1.88} & & \rebuttal{1.88} & & \rebuttal{1.63} & & \rebuttal{1.89} & & \rebuttal{\textcolor{green!50!black}{+216\%}} \\   
  & DFlash$^{\dagger}$ & 4.66 & & 5.65 & & 5.24 & & 4.64 & & 4.29 & & 4.85 & & 2.48 & & 4.55 & & \rebuttal{\textcolor{green!50!black}{+31\%}} \\
  & D-PACE             & 5.22 & & 6.41 & & 6.01 & & 5.05 & & 4.74 & & 5.35 & & 2.75 & & 5.08 & & \rebuttal{\textcolor{green!50!black}{+18\%}} \\
  & \rebuttal{\name{}} & \rebuttal{\textbf{6.36}} & & \rebuttal{\textbf{7.57}} & & \rebuttal{\textbf{6.82}} & & \rebuttal{\textbf{6.15}} & & \rebuttal{\textbf{5.65}} & & \rebuttal{\textbf{6.27}} & & \rebuttal{\textbf{3.08}} & & \rebuttal{\textbf{5.98}} & & \rebuttal{--} \\   
\midrule
\multirow{3}{*}{Qwen3-4B}
  & DFlash$^{\dagger}$ & 4.44 & & 5.43 & & 4.93 & & 4.42 & & 4.19 & & 4.35 & & 2.40 & & 4.31 & & \rebuttal{\textcolor{green!50!black}{+31\%}} \\
  & \rebuttal{D-PACE}  & \rebuttal{5.06} & & \rebuttal{5.95} & & \rebuttal{5.47} & & \rebuttal{4.74} & & \rebuttal{4.67} & & \rebuttal{4.95} & & \rebuttal{2.69} & & \rebuttal{4.79} & & \rebuttal{\textcolor{green!50!black}{+18\%}} \\
  & \rebuttal{\name{}} & \rebuttal{\textbf{6.07}} & & \rebuttal{\textbf{7.14}} & & \rebuttal{\textbf{6.42}} & & \rebuttal{\textbf{5.70}} & & \rebuttal{\textbf{5.49}} & & \rebuttal{\textbf{5.69}} & & \rebuttal{\textbf{2.96}} & & \rebuttal{\textbf{5.64}} & & \rebuttal{--} \\   
\midrule
\multirow{3}{*}{\begin{tabular}[c]{@{}l@{}}Gemma-4\\26B-A4B\end{tabular}}
  & DFlash$^{\dagger}$ & 3.20 & & 3.68 & & 3.33 & & 3.77 & & 2.70 & & 2.82 & & 1.78 & & 3.04 & & \rebuttal{\textcolor{green!50!black}{+25\%}} \\
  & \rebuttal{D-PACE}  & \rebuttal{3.40} & & \rebuttal{4.12} & & \rebuttal{3.86} & & \rebuttal{4.39} & & \rebuttal{3.02} & & \rebuttal{3.31} & & \rebuttal{1.81} & & \rebuttal{3.42} & & \rebuttal{\textcolor{green!50!black}{+11\%}} \\
  & \rebuttal{\name{}} & \rebuttal{\textbf{3.81}} & & \rebuttal{\textbf{4.60}} & & \rebuttal{\textbf{4.40}} & & \rebuttal{\textbf{4.88}} & & \rebuttal{\textbf{3.31}} & & \rebuttal{\textbf{3.72}} & & \rebuttal{\textbf{1.98}} & & \rebuttal{\textbf{3.81}} & & \rebuttal{--} \\   
\midrule
\addlinespace[1.5ex]
\multicolumn{19}{c}{Temperature = 1} \\
\midrule
\multirow{5}{*}{Qwen3-8B}
  & \rebuttal{SSD (0.6B)} & \rebuttal{2.37} & & \rebuttal{2.33} & & \rebuttal{2.14} & & \rebuttal{2.16} & & \rebuttal{2.16} & & \rebuttal{2.16} & & \rebuttal{1.78} & & \rebuttal{2.16} & & \rebuttal{\textcolor{green!50!black}{+118\%}} \\   
  & \rebuttal{SSD (1.7B)} & \rebuttal{1.93} & & \rebuttal{1.91} & & \rebuttal{1.82} & & \rebuttal{1.85} & & \rebuttal{1.82} & & \rebuttal{1.81} & & \rebuttal{1.57} & & \rebuttal{1.82} & & \rebuttal{\textcolor{green!50!black}{+158\%}} \\   
  & DFlash$^{\dagger}$ & 4.23 & & 4.56 & & 3.13 & & 3.80 & & 3.84 & & 4.46 & & 2.26 & & 3.75 & & \rebuttal{\textcolor{green!50!black}{+25\%}} \\
  & D-PACE             & 4.64 & & 4.88 & & 3.52 & & 4.09 & & 4.17 & & 4.85 & & 2.43 & & 4.08 & & \rebuttal{\textcolor{green!50!black}{+15\%}} \\
  & \rebuttal{\name{}} & \rebuttal{\textbf{5.48}} & & \rebuttal{\textbf{5.58}} & & \rebuttal{\textbf{3.97}} & & \rebuttal{\textbf{4.84}} & & \rebuttal{\textbf{4.72}} & & \rebuttal{\textbf{5.65}} & & \rebuttal{\textbf{2.68}} & & \rebuttal{\textbf{4.70}} & & \rebuttal{--} \\   
\midrule
\multirow{3}{*}{Qwen3-4B}
  & DFlash$^{\dagger}$ & 4.09 & & 4.37 & & 3.17 & & 4.01 & & 3.85 & & 3.95 & & 2.24 & & 3.67 & & \rebuttal{\textcolor{green!50!black}{+27\%}} \\
  & \rebuttal{D-PACE}  & \rebuttal{4.54} & & \rebuttal{4.69} & & \rebuttal{3.39} & & \rebuttal{4.33} & & \rebuttal{4.15} & & \rebuttal{4.58} & & \rebuttal{2.46} & & \rebuttal{4.02} & & \rebuttal{\textcolor{green!50!black}{+16\%}} \\
  & \rebuttal{\name{}} & \rebuttal{\textbf{5.41}} & & \rebuttal{\textbf{5.43}} & & \rebuttal{\textbf{3.81}} & & \rebuttal{\textbf{5.10}} & & \rebuttal{\textbf{4.93}} & & \rebuttal{\textbf{5.25}} & & \rebuttal{\textbf{2.69}} & & \rebuttal{\textbf{4.66}} & & \rebuttal{--} \\   
\midrule
\multirow{3}{*}{\begin{tabular}[c]{@{}l@{}}Gemma-4\\26B-A4B\end{tabular}}
  & DFlash$^{\dagger}$ & 3.11 & & 3.53 & & 3.00 & & 3.60 & & 2.63 & & 2.59 & & 1.74 & & 2.88 & & \rebuttal{\textcolor{green!50!black}{+19\%}} \\
  & \rebuttal{D-PACE}  & \rebuttal{3.26} & & \rebuttal{3.83} & & \rebuttal{3.31} & & \rebuttal{4.09} & & \rebuttal{2.93} & & \rebuttal{2.91} & & \rebuttal{1.76} & & \rebuttal{3.16} & & \rebuttal{\textcolor{green!50!black}{+9\%}} \\
  & \rebuttal{\name{}} & \rebuttal{\textbf{3.60}} & & \rebuttal{\textbf{4.17}} & & \rebuttal{\textbf{3.76}} & & \rebuttal{\textbf{4.39}} & & \rebuttal{\textbf{3.13}} & & \rebuttal{\textbf{3.18}} & & \rebuttal{\textbf{1.86}} & & \rebuttal{\textbf{3.44}} & & \rebuttal{--} \\   
\bottomrule
\end{tabular}}
\end{center}
\vspace{-5pt}
\end{table}

%% file: tables/vLLM_main.tex
\begin{table}[t]
\caption{
\rebuttal{
Single-request serving in vLLM on H100 GPUs with greedy decoding. We report decode throughput (tok/s) and speedup (Spd.) over AR of the same target model. 
}
}
\label{tab:vllm_serving}
\begin{center}
\small
\setlength{\tabcolsep}{4pt}
\newcommand{\gain}[1]{\textcolor{green!50!black}{+#1\%}}
\resizebox{0.95\textwidth}{!}{%
\begin{tabular*}{\linewidth}{@{\extracolsep{\fill}}ll rr rr rr rr@{}}
\toprule
\multirow{2}{*}{\bf Model} & \multirow{2}{*}{\bf Method} & \multicolumn{2}{c}{MATH-500} & \multicolumn{2}{c}{HumanEval} & \multicolumn{2}{c}{MT-Bench} & \multicolumn{2}{c}{Average} \\
\cmidrule(lr){3-4} \cmidrule(lr){5-6} \cmidrule(lr){7-8} \cmidrule(lr){9-10}
 & & tok/s & Spd. & tok/s & Spd. & tok/s & Spd. & Spd. & \name{} gain\\
\midrule
\multirow{4}{*}{Qwen3-8B}
  & AR       & 148  & 1.00 & 149  & 1.00 & 148 & 1.00 & 1.00 & -\\
  & DFlash   & 920  & 6.22 & 758  & 5.11 & 400 & 2.70 & 4.67 & \gain{24} \\
  & D-PACE   & 1046 & 7.07 & 851  & 5.73 & 442 & 2.98 & 5.26 & \gain{10} \\
  & \name{}  & \textbf{1153} & \textbf{7.79} & \textbf{944}  & \textbf{6.36} & \textbf{476} & \textbf{3.21} & \textbf{5.78} & -- \\
\midrule
\multirow{4}{*}{Qwen3-4B}
  & AR       & 241  & 1.00 & 246  & 1.00 & 241 & 1.00 & 1.00 & -\\
  & DFlash   & 1443 & 6.00 & 1234 & 5.01 & 616 & 2.56 & 4.52 & \gain{13} \\
  & D-PACE   & 1561 & 6.49 & 1323 & 5.37 & 668 & 2.77 & 4.88 & \gain{5} \\
  & \name{}  & \textbf{1670} & \textbf{6.94} & \textbf{1370} & \textbf{5.56} & \textbf{671} & \textbf{2.79} & \textbf{5.10} & -- \\
\midrule
\multirow{4}{*}{\begin{tabular}[c]{@{}l@{}}Gemma-4\\26B-A4B\end{tabular}}
  & AR       & 190  & 1.00 & 196  & 1.00 & 196 & 1.00 & 1.00 & -\\
  & DFlash   & 702  & 3.69 & 698  & 3.57 & 340 & 1.74 & 3.00 & \gain{16} \\
  & D-PACE   & 749  & 3.94 & 776  & 3.97 & 342 & 1.74 & 3.22 & \gain{8} \\
  & \name{}  & \textbf{813}  & \textbf{4.28} & \textbf{837}  & \textbf{4.28} & \textbf{366} & \textbf{1.87} & \textbf{3.48} & -- \\
\bottomrule
\end{tabular*}
}
\end{center}
\end{table}

%% file: tables/main_table_extended.tex
\begin{table}[hb]
\caption{Our main results with the mean acceptance length $\tau$ beside every speedup. All are generated with $\gamma=16$, maximum sequence length of 2048 and using the HuggingFace transformers backend with compile mode enabled. \rebuttal{SSD, reimplemented in our framework, instead uses a standalone Qwen3-0.6B or Qwen3-1.7B drafter with draft size $\gamma=2$.}}
\label{tab:main_tau}
\begin{center}
\setlength{\tabcolsep}{3pt}
\resizebox{\textwidth}{!}{%
\begin{tabular}{ll r@{}l r r@{}l r r@{}l r r@{}l r r@{}l r r@{}l r r@{}l r r@{}l r}
\toprule
\multirow{3}{*}{\bf Model} & \multirow{3}{*}{\bf Method}
  & \multicolumn{9}{c}{\textsc{Math}} & \multicolumn{9}{c}{\textsc{Code}} & \multicolumn{3}{c}{\textsc{Chat}} & \multicolumn{3}{c}{\multirow{2}{*}{\textit{Avg.}}} \\
\cmidrule(lr){3-11} \cmidrule(lr){12-20} \cmidrule(lr){21-23}
 & & \multicolumn{3}{c}{GSM8K} & \multicolumn{3}{c}{MATH-500} & \multicolumn{3}{c}{AIME26} & \multicolumn{3}{c}{HumanEval} & \multicolumn{3}{c}{MBPP} & \multicolumn{3}{c}{LCB} & \multicolumn{3}{c}{MT-Bench} & & & \\
 & & \multicolumn{2}{c}{Speedup} & $\tau$ & \multicolumn{2}{c}{Speedup} & $\tau$ & \multicolumn{2}{c}{Speedup} & $\tau$ & \multicolumn{2}{c}{Speedup} & $\tau$ & \multicolumn{2}{c}{Speedup} & $\tau$ & \multicolumn{2}{c}{Speedup} & $\tau$ & \multicolumn{2}{c}{Speedup} & $\tau$ & \multicolumn{2}{c}{Speedup} & $\tau$ \\
\midrule
\multicolumn{26}{c}{Temperature = 0} \\
\midrule
\multirow{5}{*}{Qwen3-8B}
  & \rebuttal{SSD (0.6B)} & \rebuttal{2.45} & & \rebuttal{2.72} & \rebuttal{2.45} & & \rebuttal{2.77} & \rebuttal{2.36} & & \rebuttal{2.74} & \rebuttal{2.30} & & \rebuttal{2.63} & \rebuttal{2.25} & & \rebuttal{2.54} & \rebuttal{2.30} & & \rebuttal{2.70} & \rebuttal{1.87} & & \rebuttal{2.36} & \rebuttal{2.29} & & \rebuttal{2.64} \\   
  & \rebuttal{SSD (1.7B)} & \rebuttal{2.00} & & \rebuttal{2.80} & \rebuttal{2.01} & & \rebuttal{2.84} & \rebuttal{1.94} & & \rebuttal{2.80} & \rebuttal{1.89} & & \rebuttal{2.73} & \rebuttal{1.88} & & \rebuttal{2.68} & \rebuttal{1.88} & & \rebuttal{2.76} & \rebuttal{1.63} & & \rebuttal{2.50} & \rebuttal{1.89} & & \rebuttal{2.73} \\   
  & DFlash$^{\dagger}$ & 4.66 & & 6.52 & 5.65 & & 8.14 & 5.24 & & 7.55 & 4.64 & & 6.62 & 4.29 & & 6.01 & 4.85 & & 7.30 & 2.48 & & 4.25 & 4.55 & & 6.63 \\
  & D-PACE             & 5.22 & & 7.19 & 6.41 & & 9.06 & 6.01 & & \textbf{8.55} & 5.05 & & 7.10 & 4.74 & & 6.52 & 5.35 & & 7.89 & 2.75 & & 4.65 & 5.08 & & 7.28 \\
  & \rebuttal{\name{}} & \rebuttal{\textbf{6.36}} & & \rebuttal{\textbf{7.54}} & \rebuttal{\textbf{7.57}} & & \rebuttal{\textbf{9.20}} & \rebuttal{\textbf{6.82}} & & \rebuttal{8.44} & \rebuttal{\textbf{6.15}} & & \rebuttal{\textbf{7.43}} & \rebuttal{\textbf{5.65}} & & \rebuttal{\textbf{6.73}} & \rebuttal{\textbf{6.27}} & & \rebuttal{\textbf{8.29}} & \rebuttal{\textbf{3.08}} & & \rebuttal{\textbf{4.77}} & \rebuttal{\textbf{5.98}} & & \rebuttal{\textbf{7.49}} \\   
\midrule
\multirow{3}{*}{Qwen3-4B}
  & DFlash$^{\dagger}$ & 4.44 & & 6.53 & 5.43 & & 8.23 & 4.93 & & 7.55 & 4.42 & & 6.68 & 4.19 & & 6.16 & 4.35 & & 6.91 & 2.40 & & 4.35 & 4.31 & & 6.63 \\
  & \rebuttal{D-PACE}  & \rebuttal{5.06} & & \rebuttal{7.33} & \rebuttal{5.95} & & \rebuttal{8.87} & \rebuttal{5.47} & & \rebuttal{8.24} & \rebuttal{4.74} & & \rebuttal{7.13} & \rebuttal{4.67} & & \rebuttal{6.76} & \rebuttal{4.95} & & \rebuttal{7.71} & \rebuttal{2.69} & & \rebuttal{4.80} & \rebuttal{4.79} & & \rebuttal{7.26} \\
  & \rebuttal{\name{}} & \rebuttal{\textbf{6.07}} & & \rebuttal{\textbf{7.48}} & \rebuttal{\textbf{7.14}} & & \rebuttal{\textbf{9.07}} & \rebuttal{\textbf{6.42}} & & \rebuttal{\textbf{8.51}} & \rebuttal{\textbf{5.70}} & & \rebuttal{\textbf{7.25}} & \rebuttal{\textbf{5.49}} & & \rebuttal{\textbf{6.83}} & \rebuttal{\textbf{5.69}} & & \rebuttal{\textbf{7.87}} & \rebuttal{\textbf{2.96}} & & \rebuttal{\textbf{4.86}} & \rebuttal{\textbf{5.64}} & & \rebuttal{\textbf{7.41}} \\   
\midrule
\multirow{3}{*}{\begin{tabular}[c]{@{}l@{}}Gemma-4\\26B-A4B\end{tabular}}
  & DFlash$^{\dagger}$ & 3.20 & & 6.51 & 3.68 & & 7.31 & 3.33 & & 6.52 & 3.77 & & 7.80 & 2.70 & & 5.39 & 2.82 & & 5.78 & 1.78 & & 4.19 & 3.04 & & 6.21 \\
  & \rebuttal{D-PACE}  & \rebuttal{3.40} & & \rebuttal{6.91} & \rebuttal{4.12} & & \rebuttal{8.19} & \rebuttal{3.86} & & \rebuttal{7.52} & \rebuttal{4.39} & & \rebuttal{9.28} & \rebuttal{3.02} & & \rebuttal{6.09} & \rebuttal{3.31} & & \rebuttal{6.85} & \rebuttal{1.81} & & \rebuttal{4.40} & \rebuttal{3.42} & & \rebuttal{7.03} \\
  & \rebuttal{\name{}} & \rebuttal{\textbf{3.81}} & & \rebuttal{\textbf{7.18}} & \rebuttal{\textbf{4.60}} & & \rebuttal{\textbf{8.32}} & \rebuttal{\textbf{4.40}} & & \rebuttal{\textbf{7.80}} & \rebuttal{\textbf{4.88}} & & \rebuttal{\textbf{9.44}} & \rebuttal{\textbf{3.31}} & & \rebuttal{\textbf{6.14}} & \rebuttal{\textbf{3.72}} & & \rebuttal{\textbf{6.98}} & \rebuttal{\textbf{1.98}} & & \rebuttal{\textbf{4.52}} & \rebuttal{\textbf{3.81}} & & \rebuttal{\textbf{7.20}} \\   
\midrule
\addlinespace[1.5ex]
\multicolumn{26}{c}{Temperature = 1} \\
\midrule
\multirow{5}{*}{Qwen3-8B}
  & \rebuttal{SSD (0.6B)} & \rebuttal{2.37} & & \rebuttal{2.70} & \rebuttal{2.33} & & \rebuttal{2.74} & \rebuttal{2.14} & & \rebuttal{2.66} & \rebuttal{2.16} & & \rebuttal{2.58} & \rebuttal{2.16} & & \rebuttal{2.50} & \rebuttal{2.16} & & \rebuttal{2.62} & \rebuttal{1.78} & & \rebuttal{2.32} & \rebuttal{2.16} & & \rebuttal{2.59} \\   
  & \rebuttal{SSD (1.7B)} & \rebuttal{1.93} & & \rebuttal{2.79} & \rebuttal{1.91} & & \rebuttal{2.82} & \rebuttal{1.82} & & \rebuttal{2.73} & \rebuttal{1.85} & & \rebuttal{2.71} & \rebuttal{1.82} & & \rebuttal{2.67} & \rebuttal{1.81} & & \rebuttal{2.74} & \rebuttal{1.57} & & \rebuttal{2.50} & \rebuttal{1.82} & & \rebuttal{2.71} \\   
  & DFlash$^{\dagger}$ & 4.23 & & 6.03 & 4.56 & & 6.68 & 3.13 & & 4.67 & 3.80 & & 5.42 & 3.84 & & 5.42 & 4.46 & & 6.72 & 2.26 & & 3.89 & 3.75 & & 5.55 \\
  & D-PACE             & 4.64 & & 6.59 & 4.88 & & 7.19 & 3.52 & & 5.25 & 4.09 & & 5.78 & 4.17 & & \textbf{5.82} & 4.85 & & 7.26 & 2.43 & & 4.09 & 4.08 & & 6.00 \\
  & \rebuttal{\name{}} & \rebuttal{\textbf{5.48}} & & \rebuttal{\textbf{6.77}} & \rebuttal{\textbf{5.58}} & & \rebuttal{\textbf{7.24}} & \rebuttal{\textbf{3.97}} & & \rebuttal{\textbf{5.39}} & \rebuttal{\textbf{4.84}} & & \rebuttal{\textbf{6.08}} & \rebuttal{\textbf{4.72}} & & \rebuttal{5.80} & \rebuttal{\textbf{5.65}} & & \rebuttal{\textbf{7.64}} & \rebuttal{\textbf{2.68}} & & \rebuttal{\textbf{4.21}} & \rebuttal{\textbf{4.70}} & & \rebuttal{\textbf{6.16}} \\   
\midrule
\multirow{3}{*}{Qwen3-4B}
  & DFlash$^{\dagger}$ & 4.09 & & 6.07 & 4.37 & & 6.79 & 3.17 & & 4.96 & 4.01 & & 6.05 & 3.85 & & 5.63 & 3.95 & & 6.31 & 2.24 & & 4.01 & 3.67 & & 5.69 \\
  & \rebuttal{D-PACE}  & \rebuttal{4.54} & & \rebuttal{6.64} & \rebuttal{4.69} & & \rebuttal{7.23} & \rebuttal{3.39} & & \rebuttal{5.26} & \rebuttal{4.33} & & \rebuttal{6.50} & \rebuttal{4.15} & & \rebuttal{6.06} & \rebuttal{4.58} & & \rebuttal{7.21} & \rebuttal{2.46} & & \rebuttal{\textbf{4.38}} & \rebuttal{4.02} & & \rebuttal{6.18} \\
  & \rebuttal{\name{}} & \rebuttal{\textbf{5.41}} & & \rebuttal{\textbf{6.85}} & \rebuttal{\textbf{5.43}} & & \rebuttal{\textbf{7.28}} & \rebuttal{\textbf{3.81}} & & \rebuttal{\textbf{5.45}} & \rebuttal{\textbf{5.10}} & & \rebuttal{\textbf{6.52}} & \rebuttal{\textbf{4.93}} & & \rebuttal{\textbf{6.22}} & \rebuttal{\textbf{5.25}} & & \rebuttal{\textbf{7.30}} & \rebuttal{\textbf{2.69}} & & \rebuttal{4.36} & \rebuttal{\textbf{4.66}} & & \rebuttal{\textbf{6.28}} \\   
\midrule
\multirow{3}{*}{\begin{tabular}[c]{@{}l@{}}Gemma-4\\26B-A4B\end{tabular}}
  & DFlash$^{\dagger}$ & 3.11 & & 6.32 & 3.53 & & 6.92 & 3.00 & & 5.89 & 3.60 & & 7.44 & 2.63 & & 5.24 & 2.59 & & 5.29 & 1.74 & & 4.06 & 2.88 & & 5.88 \\
  & \rebuttal{D-PACE}  & \rebuttal{3.26} & & \rebuttal{6.60} & \rebuttal{3.83} & & \rebuttal{7.61} & \rebuttal{3.31} & & \rebuttal{6.52} & \rebuttal{4.09} & & \rebuttal{8.70} & \rebuttal{2.93} & & \rebuttal{\textbf{5.88}} & \rebuttal{2.91} & & \rebuttal{6.09} & \rebuttal{1.76} & & \rebuttal{4.25} & \rebuttal{3.16} & & \rebuttal{6.52} \\
  & \rebuttal{\name{}} & \rebuttal{\textbf{3.60}} & & \rebuttal{\textbf{6.93}} & \rebuttal{\textbf{4.17}} & & \rebuttal{\textbf{7.73}} & \rebuttal{\textbf{3.76}} & & \rebuttal{\textbf{6.86}} & \rebuttal{\textbf{4.39}} & & \rebuttal{\textbf{8.81}} & \rebuttal{\textbf{3.13}} & & \rebuttal{\textbf{5.88}} & \rebuttal{\textbf{3.18}} & & \rebuttal{\textbf{6.21}} & \rebuttal{\textbf{1.86}} & & \rebuttal{\textbf{4.34}} & \rebuttal{\textbf{3.44}} & & \rebuttal{\textbf{6.68}} \\   
\bottomrule
\end{tabular}}
\end{center}
\end{table}

%% file: tables/servings_hf_vllm.tex
\begin{table}[t]
\caption{
\rebuttal{
Decoding speeds across two inference backends: vLLM and HF Transformers. \name{} on HF Transformers uses a budget of 32 while on vLLM uses 16. Greedy decoding, 128 prompts per benchmark, 2048 maximum decoded tokens per sequence.
}
}
\label{tab:serving_stacks}
\begin{center}
\small
\setlength{\tabcolsep}{4pt}
\newcommand{\gain}[1]{\textcolor{green!50!black}{+#1\%}}
\newcommand{\backend}[1]{\multicolumn{10}{l}{\textit{#1}}}
\begin{tabular*}{\linewidth}{@{\extracolsep{\fill}}ll rr rr rr rr r@{}}
\toprule
\multirow{2}{*}{\bf Model} & \multirow{2}{*}{\bf Method} & \multicolumn{2}{c}{MATH-500} & \multicolumn{2}{c}{HumanEval} & \multicolumn{2}{c}{MT-Bench} & \multicolumn{2}{c}{Average} & \multirow{2}{*}{\begin{tabular}[c]{@{}c@{}}Gain of\\\name{}\end{tabular}} \\
\cmidrule(lr){3-4} \cmidrule(lr){5-6} \cmidrule(lr){7-8} \cmidrule(lr){9-10}
 & & tok/s & $\tau$ & tok/s & $\tau$ & tok/s & $\tau$ & tok/s & $\tau$ & \\
\midrule
\multirow{9}{*}{Qwen3-8B}
  & \backend{HF Transformers (compiled)} \\
  & AR       & 128 & 1.00 & 128 & 1.00 & 129 & 1.00 & 128 & 1.00 & \\ 
  & DFlash   & 731 & 8.14 & 602 & 6.62 & 321 & 4.25 & 551 & 6.34 & \gain{30} \\ 
  & \name{}  & 970 & 9.20 & 788 & 7.43 & 396 & 4.77 & 718 & 7.13 & -- \\ 
\cmidrule(l){2-11}
  & \backend{vLLM} \\
  & AR       & 148 & 1.00 & 149 & 1.00 & 148 & 1.00 & 148 & 1.00 & \\
  & DFlash   & 920 & 8.02 & 758 & 6.49 & 400 & 4.25 & 693 & 6.25 & \gain{24} \\
  & D-PACE   & 1046 & 8.73 & 851 & 7.05 & 442 & 4.56 & 780 & 6.78 & \gain{10} \\
  & \name{}  & 1153 & 8.87 & 944 & 7.22 & 476 & 4.71 & 858 & 6.94 & -- \\
\midrule
\multirow{9}{*}{Qwen3-4B}
  & \backend{HF Transformers (compiled)} \\
  & AR       & 189 & 1.00 & 189 & 1.00 & 191 & 1.00 & 190 & 1.00 & \\ 
  & DFlash   & 1054 & 8.23 & 860 & 6.68 & 460 & 4.35 & 791 & 6.42 & \gain{28} \\ 
  & \name{}  & 1352 & 9.07 & 1078 & 7.25 & 564 & 4.86 & 998 & 7.06 & -- \\ 
\cmidrule(l){2-11}
  & \backend{vLLM} \\
  & AR       & 241 & 1.00 & 246 & 1.00 & 241 & 1.00 & 243 & 1.00 & \\
  & DFlash   & 1443 & 8.04 & 1234 & 6.74 & 616 & 4.14 & 1098 & 6.31 & \gain{13} \\
  & D-PACE   & 1561 & 8.68 & 1323 & 7.19 & 668 & 4.56 & 1184 & 6.81 & \gain{5} \\
  & \name{}  & 1670 & 8.79 & 1370 & 7.23 & 671 & 4.58 & 1237 & 6.87 & -- \\
\midrule
\multirow{9}{*}{\begin{tabular}[c]{@{}l@{}}Gemma-4\\26B-A4B\end{tabular}}
  & \backend{HF Transformers (compiled)} \\
  & AR       & 154 & 1.00 & 155 & 1.00 & 156 & 1.00 & 155 & 1.00 & \\ 
  & DFlash   & 581 & 7.31 & 594 & 7.80 & 280 & 4.19 & 485 & 6.43 & \gain{22} \\ 
  & \name{}  & 709 & 8.32 & 757 & 9.44 & 308 & 4.52 & 591 & 7.43 & -- \\ 
\cmidrule(l){2-11}
  & \backend{vLLM} \\
  & AR       & 190 & 1.00 & 196 & 1.00 & 196 & 1.00 & 194 & 1.00 & \\
  & DFlash   & 702 & 7.37 & 698 & 8.31 & 340 & 4.56 & 580 & 6.75 & \gain{16} \\
  & D-PACE   & 749 & 7.84 & 776 & 9.65 & 342 & 4.60 & 622 & 7.36 & \gain{8} \\
  & \name{}  & 813 & 7.93 & 837 & 9.76 & 366 & 4.60 & 672 & 7.43 & -- \\
\bottomrule
\end{tabular*}
\end{center}
\end{table}

%% file: tables/serving_concurrent.tex
\begin{table}[t]
\caption{
\rebuttal{
Concurrent serving with vLLM on Gemma-4-26B-A4B (SPEED-Bench throughput splits, 1536 requests each, greedy, H100, vLLM 0.31). D-PACE runs on vLLM's built-in DFlash at block 16 and 8; \name{} serves the anchor-free drafter through our plugin at block 8, the drafter on a second GPU, from the placeholder token $[P]$ without and with the Markov head ($r=64$, $[P]$+head). Output throughput over all requests at concurrency $c$, input length 1k and 2k tokens; the best per column in bold, the second best underlined. Gain: relative throughput improvement of \name{} ($[P]$+head) over the faster D-PACE configuration.
}
}
\label{tab:serving_final}
\begin{center}
\small
\setlength{\tabcolsep}{3pt}
\begin{tabular}{ll rrrrrrr}
\toprule
\multirow{2}{*}{\bf Input} & \multirow{2}{*}{\bf Method} & \multicolumn{7}{c}{Output tok/s at concurrency $c$} \\
\cmidrule(lr){3-9}
 & & 1 & 2 & 4 & 8 & 16 & 32 & 64 \\
\midrule
\multirow{6}{*}{1k}
  & AR                           & 174 & 280 & 441 & 702 & 1083 & 1597 & 2365 \\
  & D-PACE (16)                  & 270 & 442 & 716 & 1156 & 1878 & 2871 & 3037 \\
  & D-PACE (8)                   & 284 & 459 & 739 & \underline{1198} & \underline{1920} & \textbf{3113} & \underline{4132} \\
  & \name{} ($[P]$)              & \underline{298} & \underline{475} & \underline{747} & 1183 & 1899 & 2943 & 3957 \\
  & \name{} ($[P]$+head)         & \textbf{308} & \textbf{487} & \textbf{769} & \textbf{1233} & \textbf{1953} & \underline{3066} & \textbf{4175} \\
\cmidrule(lr){2-9}
  & Gain of \name{}              & \textcolor{green!50!black}{+8.2\%} & \textcolor{green!50!black}{+6.2\%} & \textcolor{green!50!black}{+4.1\%} & \textcolor{green!50!black}{+2.9\%} & \textcolor{green!50!black}{+1.7\%} & \textcolor{red!70!black}{$-$1.5\%} & \textcolor{green!50!black}{+1.0\%} \\
\midrule
\multirow{6}{*}{2k}
  & AR                           & 164 & 278 & 426 & 694 & 1033 & 1562 & 2315 \\
  & D-PACE (16)                  & 263 & 427 & 686 & 1104 & 1802 & 2686 & 2580 \\
  & D-PACE (8)                   & 272 & 445 & 704 & 1130 & 1868 & \textbf{2826} & \underline{3653} \\
  & \name{} ($[P]$)              & \underline{291} & \underline{467} & \underline{738} & \underline{1189} & \underline{1874} & 2681 & 3604 \\
  & \name{} ($[P]$+head)         & \textbf{302} & \textbf{480} & \textbf{763} & \textbf{1219} & \textbf{1918} & \underline{2761} & \textbf{3750} \\
\cmidrule(lr){2-9}
  & Gain of \name{}              & \textcolor{green!50!black}{+11.1\%} & \textcolor{green!50!black}{+8.0\%} & \textcolor{green!50!black}{+8.3\%} & \textcolor{green!50!black}{+7.9\%} & \textcolor{green!50!black}{+2.7\%} & \textcolor{red!70!black}{$-$2.3\%} & \textcolor{green!50!black}{+2.7\%} \\
\bottomrule
\end{tabular}
\end{center}
\end{table}

%% file: tables/cpu.tex
\begin{table}[t]
\caption{
\rebuttal{
Drafting on the CPU for Gemma-4-26B-A4B-it (greedy, $\gamma=16$, 2048 new tokens, Hugging Face Transformers in compiled mode): speedup over autoregressive decoding, mean acceptance length, and cache hit rate. D-PACE and the CPU drafter are measured in the same run on one node; the 2-GPU \name{} row corresponds to the run in Table~\ref{tab:main}. 
}
}
\label{tab:cpu_drafter}
\begin{center}
\setlength{\tabcolsep}{4pt}
\resizebox{\textwidth}{!}{%
\begin{tabular}{l r r r r r r r r c}
\toprule
\multirow{2}{*}{\bf Method} & \multicolumn{3}{c}{\textsc{Math}} & \multicolumn{3}{c}{\textsc{Code}} & \multicolumn{1}{c}{\textsc{Chat}} & \multirow{2}{*}{\textit{Avg.}} & \multirow{2}{*}{\begin{tabular}[c]{@{}c@{}}Gain over\\D-PACE\end{tabular}} \\
\cmidrule(lr){2-4} \cmidrule(lr){5-7} \cmidrule(lr){8-8}
& GSM8K & MATH-500 & AIME26 & HumanEval & MBPP & LCB & MT-Bench & & \\
\midrule
\multicolumn{10}{l}{\textit{Speedup over AR}} \\
  D-PACE & 3.38 & 4.12 & 3.85 & 4.30 & 3.00 & 3.26 & 1.80 & 3.39 & -- \\
  \name{} (2 GPUs, B=32) & 3.81 & 4.60 & 4.40 & 4.88 & 3.31 & 3.72 & 1.98 & 3.81 & \textcolor{green!50!black}{+12.5\%} \\
  \name{} (CPU drafter, B=4) & 3.52 & 4.26 & 3.95 & 4.46 & 3.09 & 3.34 & 1.83 & 3.49 & \textcolor{green!50!black}{+3.1\%} \\
\midrule
\multicolumn{10}{l}{\textit{Acceptance length $\tau$}} \\
  D-PACE & 6.95 & 8.21 & 7.49 & 9.16 & 6.07 & 6.74 & 4.40 & 7.00 & -- \\
  \name{} (2 GPUs, B=32) & 7.18 & 8.32 & 7.80 & 9.44 & 6.14 & 6.98 & 4.52 & 7.20 & \textcolor{green!50!black}{+2.8\%} \\
  \name{} (CPU drafter, B=4) & 7.08 & 8.30 & 7.53 & 9.23 & 6.12 & 6.82 & 4.45 & 7.08 & \textcolor{green!50!black}{+1.0\%} \\
\midrule
\multicolumn{10}{l}{\textit{Cache hit rate (\%)}} \\
  \name{} (2 GPUs, B=32) & 83.5 & 89.6 & 89.0 & 88.7 & 86.3 & 85.0 & 75.3 & 85.3 & -- \\
  \name{} (CPU drafter, B=4) & 51.0 & 59.3 & 58.2 & 60.1 & 52.8 & 52.7 & 42.0 & 53.7 & -- \\
\bottomrule
\end{tabular}}
\end{center}
\end{table}

%% file: tables/ablation_cache_loop.tex
\begin{table*}[t]
\centering
\small
\caption{Miss rate (MR, \%: the share of rejected blocks whose correction was not cached) and speedup ($\times$) over autoregressive generation by number of loops, strategy, and budget. The target model is Qwen3-8B, the drafter is our \name{} model from Table \ref{tab:main}. Each value is an average over Math500, HumanEval and MT-Bench.}
\label{tab:ablation-cache-loop}
\setlength{\tabcolsep}{4pt}
\begin{tabular}{ll*{10}{c}}
\toprule
 & & \multicolumn{2}{c}{Budget 8} & \multicolumn{2}{c}{Budget 16} & \multicolumn{2}{c}{Budget 32} & \multicolumn{2}{c}{Budget 64} & \multicolumn{2}{c}{Budget 128} \\
\cmidrule(lr){3-4} \cmidrule(lr){5-6} \cmidrule(lr){7-8} \cmidrule(lr){9-10} \cmidrule(lr){11-12}
Loops & Strategy & MR & Speedup & MR & Speedup & MR & Speedup & MR & Speedup & MR & Speedup \\
\midrule
1 & Uniform   & 61.6 & 4.62 & 51.0 & 4.75 & 34.7 & 4.83 & 21.8 & 4.39 & 12.9 & 3.38 \\
  & Geometric & 62.0 & 4.70 & 46.4 & 4.78 & 30.1 & 4.85 & 18.4 & 4.40 & 10.6 & 3.39 \\
  & Dynamic   & 36.5 & 4.93 & 24.5 & 4.97 & 15.3 & 5.05 & 9.2 & \textbf{4.50} & \textbf{5.2} & \textbf{3.43} \\
\midrule
3 & Uniform   & 61.2 & 4.81 & 50.4 & 5.08 & 33.8 & 5.26 & 20.6 & 4.00 & 12.2 & 2.27 \\
  & Geometric & 61.5 & 4.97 & 45.7 & 5.18 & 29.3 & 5.28 & 17.4 & 3.99 & 9.9 & 2.27 \\
  & Dynamic   & \textbf{35.8} & \textbf{5.34} & \textbf{23.6} & \textbf{5.49} & \textbf{14.7} & \textbf{5.60} & \textbf{8.9} & 4.08 & \textbf{5.2} & 2.32 \\
\bottomrule
\end{tabular}
\end{table*}

%% file: tables/loopvsdeep.tex
\begin{table}[t]
\caption{
\rebuttal{
Looping versus depth. \name{} (five layers, the middle block looped). Speedup over AR decoding; Qwen3-8B, greedy, mean over MATH500, HumanEval, and MTBench.
}
}
\label{tab:depth_vs_loop}
\begin{center}
\resizebox{\textwidth}{!}{%
\begin{tabular}{lccccccccc}
\toprule
\multirow{2}{*}{\bf Drafter} & \multirow{2}{*}{\bf Draft layers} & \multirow{2}{*}{\bf Params (B)} & \multicolumn{4}{c}{\bf Macro-sequential drafting} & \multicolumn{3}{c}{\bf Macro-parallel drafting} \\
\cmidrule(lr){4-7} \cmidrule(lr){8-10}
 & & & Passes/draft & $\tau$ & Draft (ms/step) & Speedup & $\tau$ & Miss draft (ms) & Speedup \\
\midrule
Default (1 loop)  & 5 & 1.05 & 5 & 5.45 & 2.23 & \textbf{3.53$\times$} & 5.44 & 1.88 & 4.20$\times$ \\
Default (2 loops) & 5 & 1.05 & 8 & 5.51 & 3.00 & 3.32$\times$ & 5.52 & 1.88 & 4.24$\times$ \\
8-layer    & 8 & 1.63 & 8 & 5.53 & 2.95 & 3.39$\times$ & 5.52 & 2.61 & 4.23$\times$ \\
Default  (3 loops) & 5 & 1.05 & 11 & \textbf{5.70} & 3.77 & 3.24$\times$ & \textbf{5.66} & 1.88 & \textbf{4.36$\times$} \\
11-layer  & 11 & 2.21 & 11 & 5.62 & 3.75 & 3.22$\times$ & 5.62 & 3.36 & 4.26$\times$ \\
\bottomrule
\end{tabular}
}
\end{center}
\end{table}

%% file: tables/architectural_params.tex
\begin{table*}[t]
\centering
\small
\setlength{\tabcolsep}{5pt}
\resizebox{\textwidth}{!}{%
\begin{tabular}{l cc cc cc}
\toprule
 & \multicolumn{2}{c}{\textbf{Qwen3-4B}} & \multicolumn{2}{c}{\textbf{Qwen3-8B}} & \multicolumn{2}{c}{\textbf{Gemma-4-26B-A4B}} \\
\cmidrule(lr){2-3} \cmidrule(lr){4-5} \cmidrule(lr){6-7}
 & Target & Drafter & Target & Drafter & Target & Drafter \\
\midrule
Parameters        & 4.0B & 0.54B & 8.2B & 1.05B & 25.8B (4B act.) & 0.42B \\
Layers                        & 36   & 5 & 36 & 5 & 30 & 5 \\
Hidden size                   & 2560 & 2560 & 4096 & 4096 & 2816 & 2816 \\
FFN width                     & 9728 & 9728 & 12288 & 12288 & 704$\times$128 exp.\ (top-8) & 5632 \\
FFN type                      & SwiGLU & SwiGLU & SwiGLU & SwiGLU & MoE, GeGLU & SwiGLU \\
Attention heads (Q / KV)      & 32 / 8 & 32 / 8 & 32 / 8 & 32 / 8 & 16 / 8 (global: 16 / 2) & 32 / 8 \\
Head dimension                & 128 & 128 & 128 & 128 & 256 (global: 512) & 128 \\
Vocabulary                    & \multicolumn{2}{c}{151{,}936 (shared)} & \multicolumn{2}{c}{151{,}936 (shared)} & \multicolumn{2}{c}{262{,}144 (shared)} \\
\addlinespace
$\{l_1 \cdots l_m\}$          & --- & 1, 9, 17, 25, 33 & --- & 1, 9, 17, 25, 33 & --- & 6, 11, 17, 22, 27 \\
Draft block size $\gamma$          & --- & 16 & --- & 16 & --- & 16 \\
\bottomrule
\end{tabular}}
\caption{Architectural hyperparameters for the three target models and the DFlash drafters. }
\label{tab:architectural-params}
\end{table*}

%% file: listings/communication_logic.tex
\lstdefinestyle{ssd}{
  language=Python,
  basicstyle=\ttfamily\footnotesize,
  commentstyle=\color{gray}\itshape,
  keywordstyle=\bfseries,
  numbers=left, numberstyle=\tiny\color{gray}, numbersep=6pt,
  columns=fullflexible, keepspaces=true,
  frame=single, framesep=4pt, xleftmargin=16pt,
  showstringspaces=false, breaklines=true,
  mathescape=true,
  emph={isend,irecv,send,recv,wait},
  emphstyle=\color{blue!60!black}\bfseries,
}

\begin{lstlisting}[style=ssd,
  caption={Verifier rank: one decoding step. The $m$ selected layers
  $l_1,\dots,l_m$ cut the verifier into $m{+}1$ CUDA graphs; each latent
  $h^{(l_k)}$ is sent non-blocking as soon as its segment has been replayed.},
  label={lst:ssd-verifier}]
link.send([$\hat{\tau}$, $x_{n+1}$])                   # request
$\hat{x}_{n+2:n+\gamma}$ = link.recv()                  # reply: cached or jit draft
for $k$ in range(1, $m$ + 2):
    $h$ = graphs[$k$].replay()                  # layers $l_{k-1}{+}1 \dots l_k$
    if $k$ <= $m$:
        link.isend($h$)                        # latent $h^{(l_k)} \in \mathbb{R}^{\gamma \times d_T}$
verify($\mathrm{LM}_{\text{head}}$($h$), $\hat{x}_{n+2:n+\gamma}$)
\end{lstlisting}

\begin{lstlisting}[style=ssd,
  caption={Drafter rank: one decoding step. The receive of each latent is
  posted right before the drafter layer that consumes it; \texttt{wait()} is a
  stream-level wait, so $\mathcal{D}^{(k)}$ runs while the verifier computes
  $h^{(l_{k+1})}$.},
  label={lst:ssd-drafter}]
$\hat{\tau}$, $x_{n+1}$ = link.recv()                   # request
link.send(lookup_or_jit_draft($\hat{\tau}$, $x_{n+1}$))  # reply
for $k$ in range(1, $m$ + 1):
    $h^{(l_k)}$ = link.irecv().wait()            # posted only now
    $\mathcal{D}^{(k)}$($h^{(l_k)}$)                        # branch drafts, overlapped
\end{lstlisting}

%% file: listings/conditioning.tex
\begin{lstlisting}[style=ssd,
  caption={Prefix and depth-causal conditioning for drafter layer
  $\mathcal{D}^{(l)}$. $a, b \in \mathbb{R}^{m}$ are the learned mixing logits,
  shared by every drafter layer. The $-\infty$ mask makes $\beta_{i,l} = 0$
  (weight and gradient) for every selected layer $l_i$ not yet computed, and
  renormalises the remaining $i \le l$; this only rescales them by a per-layer
  constant, which RMSNorm removes, so $\beta_{\cdot,l}$ acts as the shared $\beta$
  restricted to $i \le l$.},
  label={lst:ssd-conditioning}]
$u_{t,i}$ = $W^{(i)} h^{(l_i)}_t$                  # projected latents, $i = 1..m$
$\alpha$      = softmax($a$)                           # all $m$ latents, same for every $l$
$\beta_{\cdot,l}$  = softmax($b$.masked_fill($i > l$, $-\infty$))    # depth-causal: $i \le l$ only
for $t$ in $1..n$:                        # prefix (previously verified)
    $c^{(l)}_t$ = RMSNorm($\sum_{i=1}^{m} \alpha_{i}\, u_{t,i}$)
for $t$ in $n{+}1..n{+}\hat{\tau}{-}1$:                # block under verification
    $c^{(l)}_t$ = RMSNorm($\sum_{i=1}^{l} \beta_{i,l}\, u_{t,i}$)
\end{lstlisting}